\documentclass{article}

\usepackage[nonatbib,preprint]{neurips_2025}
\usepackage[numbers]{natbib} 
\usepackage{tablefootnote}
 
\usepackage{scalefnt,letltxmacro,tabularx}
\usepackage[acronym,nowarn]{glossaries}
\glsdisablehyper
\makeglossaries
\usepackage{xspace}
\usepackage{float}
\usepackage{physics}
\usepackage[normalem]{ulem}
\usepackage[english]{babel}

\usepackage{enumitem} 
\usepackage{booktabs} 
\usepackage{array}    
\usepackage{wrapfig}  
\usepackage{lipsum}   
\usepackage[]{microtype} 
\usepackage{fancyvrb}
\usepackage{tabularx}

\usepackage{amssymb}
\usepackage{amsmath}
\usepackage{amsfonts}
\usepackage{mathtools}

\usepackage[dvipsnames,svgnames,table]{xcolor}
\usepackage[colorlinks]{hyperref}
\definecolor{pink3}{HTML}{fc0557}
\definecolor{shadecolor}{gray}{0.9}
\definecolor{mylightgray}{gray}{0.94}
\usepackage[all]{hypcap}
\hypersetup{
  citecolor=pink3,
  linkcolor=pink3,
  urlcolor=pink3
}
\usepackage[most,skins,theorems]{tcolorbox}

\usepackage[capitalize,nameinlink]{cleveref}
\crefname{section}{\S}{\S\S}
\Crefname{section}{\S}{\S\S}

\usepackage{graphicx}
\usepackage{subcaption}
\usepackage{tikz}
\usetikzlibrary{shapes.geometric}
\usepackage[export]{adjustbox}
\usepackage{pgf-pie}

\usepackage{pgfplots}
\pgfplotsset{compat=1.6}
\usepgfplotslibrary{groupplots}
\pgfplotsset{
  tick label style = {font=\sffamily},
  every axis label/.append style={font=\sffamily},
  typeset ticklabels with strut,
  xticklabel={$\mathsf{\pgfmathprintnumber{\tick}}$},
  yticklabel={$\mathsf{\pgfmathprintnumber{\tick}}$},
  every axis title/.append style={yshift=-1ex}
}

\usepackage[colorinlistoftodos,
  textsize=scriptsize,
  linecolor=BurntOrange,
  bordercolor=BurntOrange,
  backgroundcolor=BurntOrange!25]{todonotes}

\usepackage{multirow}
\usepackage{pifont}

\newcommand{\xmark}{\ding{55}}

\usepackage{url}

\tcbset{
  promptstyle/.style={
      enhanced,
      colback=white,
      colframe=black,
      colbacktitle=gray!20,
      width=0.98\linewidth,
      coltitle=black,
      rounded corners,
      sharp corners=north,
      boxrule=0.5pt,
      drop shadow=black!50!white,
      attach boxed title to top left={
          xshift=-2mm,
          yshift=-2mm
        },
      title code={%
          \begin{minipage}{0.8\linewidth}
            \raggedright\thetitle
          \end{minipage}
        },
      boxed title style={
          rounded corners,
          size=small,
          colback=gray!20,
        },
      fonttitle=\normalsize\bfseries,
      before upper={\parindent15pt\raggedright}
    }
}

\tcbset{
  aibox/.style={
      width=\linewidth,
      top=8pt,
      bottom=4pt,
      colback=blue!6!white,
      colframe=black,
      colbacktitle=black,
      enhanced,
      center,
      attach boxed title to top left={yshift=-0.1in,xshift=0.15in},
      boxed title style={boxrule=0pt,colframe=white,},
    }
}
\newtcolorbox{AIbox}[2][]{aibox,title=#2,#1}

\usepackage{listings}
\lstdefinestyle{TinyJSON}{
morekeywords={
    _question_, _evidence_, _database_schema_, im_start, im_end, user, system, assistant, <answer>, think, Wait,
  },
basicstyle=\ttfamily\footnotesize, 
keywordstyle=\color{purple}\bfseries, 
stringstyle=\color{purple}, 
commentstyle=\color{gray}\itshape, 
numbers=none, 
numberstyle=\tiny\color{gray}, 
stepnumber=1, 
numbersep=10pt, 
tabsize=2, 
captionpos=b, 
breaklines=true, 
breakatwhitespace=true, 
showspaces=false, 
showstringspaces=false, 
escapeinside={(*@}{@*)}, 
frame=none, 
extendedchars=true,
literate={á}{{\'a}}1 {ã}{{\~a}}1 {é}{{\'e}}1 {í}{{\'i}}1 {ó}{{\'o}}1 {ú}{{\'u}}1 {ç}{{\c{c}}}1,
}

\newacronym{SGD}{SGD}{stochastic gradient descent}
\newacronym{SGA}{SGA}{stochastic gradient ascent}
\newacronym{MAP}{MAP}{maximum-a-posteriori}
\newacronym{MLE}{MLE}{maximum likelihood estimation}
\newacronym{MNLL}{MNLL}{mean negative log-likelihood}
\newacronym{NLL}{NLL}{negative log-likelihood}
\newacronym{LL}{LL}{log-likelihood}
\newacronym{RMSE}{RMSE}{root mean square error}
\newacronym{ECE}{ECE}{expected calibration error}
\newacronym{SNR}{SNR}{signal-to-noise ratio}
\newacronym{FID}{FID}{Fr\'echet Inception Distance}
\newacronym{BPD}{BPD}{bit per dimension}
\newacronym{NFE}{NFE}{neural function evaluations}

\newacronym{AE}{AE}{autoencoder}
\newacronym{WAE}{WAE}{Wasserstein Autoencoder}
\newacronym{VAE}{VAE}{Variational Autoencoder}
\newacronym{BAE}{BAE}{Bayesian autoencoder}
\newacronym{GAN}{GAN}{Generative Adversarial Network}
\newacronym{DPGMM}{DPGMM}{Dirichlet process Gaussian mixture model}

\newacronym{MC}{MC}{Monte Carlo}
\newacronym{MCMC}{MCMC}{Markov chain Monte Carlo}
\newacronym{HMC}{HMC}{Hamiltonian Monte Carlo}
\newacronym{MH}{MH}{Metropolis-Hastings}
\newacronym{NUTS}{NUTS}{no-u-turn sampler}
\newacronym{SGHMC}{SGHMC}{stochastic gradient Hamiltonian Monte Carlo}
\newacronym{OU}{OU}{Ornstein-Uhlenbeck}

\newacronym{SDE}{SDE}{Stochastic Differential Equation}
\newacronym{CNF}{CNF}{Continuous Normalizing Flow}
\newacronym{ODE}{ODE}{Ordinary Differential Equation}
\newacronym{NF}{NF}{normalizing flow}

\newacronym{GP}{GP}{Gaussian Process}
\newacronym[longplural=deep Gaussian processes]{DGP}{DGP}{deep Gaussian process}
\newacronym{GPLVM}{GPLVM}{Gaussian process latent variable model}
\newacronym{DPMM}{DPMM}{Dirichlet Process Mixture Model}

\newacronym{LLM}{LLM}{large language model}
\newacronym{VI}{VI}{variational inference}
\newacronym{SVI}{SVI}{stochastic variational inference}
\newacronym{BNN}{BNN}{Bayesian neural network}
\newacronym{DNN}{DNN}{deep neural network}
\newacronym{CNN}{CNN}{convolutional neural network}
\newacronym{MLP}{MLP}{multilayer perceptron}
\newacronym{NN}{NN}{neural network}
\newacronym{RELU}{ReLU}{rectified linear unit}

\newacronym{ELBO}{ELBO}{evidence lower bound}
\newacronym{NELBO}{NELBO}{negative evidence lower bound}
\newacronym{ELL}{ELL}{expected log likelihood}
\newacronym{KL}{KL}{Kullback-Leibler divergence}
\newacronym{AUC}{AUC}{area under the curve}
\newacronym{VFE}{VFE}{variational free energy}

\newacronym{RBF}{RBF}{radial basis function}
\newacronym{ARD}{ARD}{automatic relevance determination}
\newacronym{RKHS}{RKHS}{reproducing kernel Hilbert space}

\newacronym{OT}{OT}{optimal transport}
\newacronym{WD}{WD}{Wasserstein distance}
\newacronym{SWD}{SWD}{sliced-Wasserstein distance}
\newacronym{DSWD}{DSWD}{distributional sliced-Wasserstein distance}

\newacronym{LAP}{LAP}{linear assignment problem}
\newacronym{SOLAP}{SOLAP}{sum of bilinear assignment problems}

\newacronym{ICL}{ICL}{in-context learning}
\newacronym{LoRA}{LoRA}{Low-Rank Adaptation}
\newacronym{SFT}{SFT}{Supervised Fine-Tuning}
\newacronym{RL}{RL}{reinforcement learning}
\newacronym{RLHF}{RLHF}{reinforcement learning from human feedback}
\newacronym{PPO}{PPO}{Proximal Policy Optimization}
\newacronym{GRPO}{GRPO}{Grouped Proximal Policy Optimization}
\newacronym{COT}{CoT}{chain-of-thought}

\usepackage[hyperpageref]{backref}
\renewcommand*{\backrefalt}[4]{%
\ifcase #1 %
No citations.%
\or
(p. #2)%
\else
(pp. #2)%
\fi
}

\newcommand{\stitle}[1]{\vspace{1ex}\noindent{\textbf{#1}}}

\title{RelationalFactQA: A Benchmark for Evaluating Tabular Fact Retrieval from Large Language Models}

\author{%
    Dario Satriani, Enzo Veltri, Donatello Santoro\\
    University of Basilicata, Potenza, Italy\\
   \texttt{name.surname@unibas.it}\\
   \And
   \And
   Paolo Papotti \\
   EURECOM, Biot, France \\
   \texttt{paolo.papotti@eurecom.fr}\\
}

\begin{document}

\maketitle

\begin{abstract}
 
Factuality in Large Language Models (LLMs) is a persistent challenge. Current benchmarks often assess short factual answers, overlooking the critical ability to generate structured, multi-record tabular outputs from parametric knowledge. We demonstrate that this relational fact retrieval is substantially more difficult than isolated point-wise queries, even when individual facts are known to the model, exposing distinct failure modes sensitive to output dimensionality (e.g., number of attributes or records). To systematically evaluate this under-explored capability, we introduce RelationalFactQA, a new benchmark featuring diverse natural language questions (paired with SQL) and gold-standard tabular answers, specifically designed to assess knowledge retrieval in a structured format. RelationalFactQA enables analysis across varying query complexities, output sizes, and data characteristics. Our experiments reveal that even state-of-the-art LLMs struggle significantly, not exceeding 25\% factual accuracy in generating relational outputs, with performance notably degrading as output dimensionality increases. These findings underscore critical limitations in current LLMs' ability to synthesize structured factual knowledge and establish RelationalFactQA as a crucial resource for measuring future progress in LLM factuality.
\end{abstract}



\section{Introduction}

Large Language Models (LLMs) have emerged as powerful tools capable of understanding and generating human-like text. Despite these advances, \textit{factuality} -- the ability of LLMs to provide responses that are truthful and faithful to the real-world knowledge encountered during pre-training -- remains a persistent challenge \cite{Ji2023SurveyHallucination,openai2023gpt4}. 
Effectively, a lack of factuality manifests as `hallucination' — the generation of plausible yet incorrect information — a pervasive issue that is still observed in frontier models~\cite{chowdhury2025truthfulness,hallReport}. 
This issue is particularly critical when LLMs are used in settings demanding high factual precision, such as medical information synthesis \cite{Singhal2023LargeLanguageModelsMedicine}, financial reporting \cite{DONG2024100715}, scientific data analysis \cite{truhn2023large}, or educational content generation \cite{Kasneci2023ChatGPTEducation}.

To evaluate and improve factual performance, the research community has developed a variety of benchmarks. However, existing benchmarks predominantly focus on single-value factuality, where the expected output is a short text span or a single scalar value (e.g., a date or named entity, or a numerical value)~\cite{simpleQAOpenAI}. These tasks often emphasize reasoning complexity (e.g., multi-hop QA or ambiguous phrasing)~\cite{li2024bird, wu2025mmqa, yu2018spider} but overlook a fundamental aspect of factual competence: the ability of LLMs to generate long, coherent  outputs directly from their internal parametric knowledge (i.e., the facts stored implicitly within the model's parameters), without retrieving external documents.

In this work, we focus on structured, multi-record, tabular outputs to investigate the factuality of LLMs in synthesizing long sequences of facts. This task is motivated by two main arguments.

\textbf{Output Size Matters.} First, experiments highlight that retrieving tabular data from parametric memory presents a significantly greater challenge than recalling isolated cell values, even when the underlying facts are known to the model. For instance, prompting an LLM to return two attributes (e.g., name and state) for US counties yields near-perfect results. However, requesting additional attributes for the same set of counties (e.g., including county area) introduces factual errors in the results. 
\begin{wrapfigure}{r}{0.42\textwidth}   
\vspace{-\baselineskip}                 

\centering
\definecolor{darkgreen}{rgb}{0,0.45,0}
\begin{tikzpicture}
\node[inner sep=0pt] {%
{\small     
\begin{tabular}{|c|c|c|}
\hline
\textbf{County} & \textbf{State} & \textbf{Area (sq mi)} \\ \hline
Los Angeles & California & 4\,751 \\ \hline
Cook        & Illinois    & 1\,635 \\ \hline
Maricopa    & Arizona     & {\color{red}8\,500} \xmark \\ \hline
\end{tabular}}};
\end{tikzpicture}
\small
\textbf{Q}: What is the area of Maricopa county?\\
\textbf{A}: \textcolor{darkgreen}{9\,224 sq mi} \checkmark\\[-0.5em] 
\end{wrapfigure}
Crucially, if we then query the LLM for these specific incorrectly reported values in isolation (e.g., ``What is the area of Maricopa county?"), the model returns the correct value, demonstrating that the error lies in the generation process, not in the absence of the underlying factual knowledge. Results show that the accuracy of retrieving a specific attribute (e.g., state) degrades linearly (from ~1.0 to ~0.2) as the total number of concurrently requested attributes increases from one to fifty, regardless of the target attribute's position in the schema. These findings underscore that the structured, multi-attribute retrieval of factual data is not merely an extension of single-fact recall but a distinct capability with unique failure modes.
Moreover, while it is hard to quantify precisely and with fine granularity the quality of the output in unstructured generation tasks~\cite{coli_a_00561}, structured data allows punctual comparison at the single fact (cell) level.

{\bf Increasing Importance of Tabular Output.} Second, we argue that the \textit{structured factual retrieval} capability of LLMs is both under-explored and essential. Several tasks require not just isolated facts about common world knowledge, but the generation of relational data: lists of entities, comparisons, and collections of items satisfying specific conditions \cite{lotus,palimpzestCIDR,docetl,galois1}. 
This requirement has been reported in sociology~\cite{00491241251336794}, business use cases~\cite{zhang2025tablellmenablingtabulardata}, medical diagnosis~\cite{bisercic2023interpretablemedicaldiagnosticsstructured}, and financial use cases~\cite{computers13100257}. 
Obtaining tabular data is also increasingly relevant for user-facing applications, such as generating comparative tables of e-commerce products or structuring personalized trip itineraries~\cite{frai.2025.1558938,tang-etal-2024-itinera}. 
Yet, current benchmarks fall short in measuring this dimension. Existing datasets that do contain tabular data focus on its role as contextual input to the LLM, in the role of a corpus for question answering or fact checking~\citep{Chen2020TabFact, Pasupat2015WikiTableQuestions, Zhong2017Seq2SQL, Aly2021FEVEROUS}.

We define the \textit{Relational Fact Retrieval} task as follows: given a query, the LLM must generate a structured table (rows and columns) containing factual information drawn purely from its parametric memory, prohibiting the use of external tools like web browsers during generation. To address the need for evaluating this capability, we introduce \textbf{RelationalFactQA}, a new benchmark designed to test LLMs’ ability to return factual knowledge in relational (i.e., tabular) form in a \textit{closed-books} setting. 
RelationalFactQA probes this capability across several dimensions. 
The benchmark contains triples with the {natural language (NL) question, the corresponding SQL script, and the expected answer in tabular format. For the creation, we combine manually crafted questions (for linguistic variety) with systematically generated ones where the corresponding query complexity (e.g., specific SQL constructs) is controlled. Expected output tables spans from small ones, with few tuples and attributes, to large ones.
These dimensions enable analysis of LLMs' performance across different logical operations (e.g., aggregates, filtering), data types (e.g., numerical, categorical), and retrieval methods (prompts with NL questions vs SQL queries). 

Through extensive experimentation, we find that although larger models show improvement, the ability to produce correct structured answers remains limited — especially as the number of tuples and attributes increases or the query involves less common facts and numerical conditions. 
Moreover, we observe that even state-of-the-art models rarely exceed 25\% of factual accuracy on our benchmark.

To summarize, this paper makes the following contributions:

\begin{itemize}[leftmargin=*]
\item \textbf{Task formulation.}  We introduce \emph{Relational Fact Retrieval} — the closed-book generation of multi-tuple, multi-attribute tables directly from an LLM’s parametric memory — and clarify how it differs from single-fact recall and context-based table QA.  

\item \textbf{RelationalFactQA benchmark.} We release a 696-question dataset covering nine knowledge domains, each triple-annotated with a natural-language query, its equivalent SQL statement, and a fully verified gold table (avg.\ 27 rows $\times$ 5 attrs).  

\item \textbf{Hybrid construction pipeline.}  Our semi-automatic workflow unifies (i) manual curation from three existing corpora and (ii) YAGO-driven synthetic tables, yielding controlled variation in schema size, output size, and query complexity.  


\item \textbf{Comprehensive empirical study.}  Nine  LLMs (7B – 235B params) are benchmarked under three retrieval techniques (NL, SQL, Chain-of-Thought). Despite parameter scaling, no model exceeds 0.25 in tuple accuracy; performance degrades linearly with requested attributes. 
Code, prompts, and data will be open-sourced to drive future progress.
\end{itemize}

Our findings lay the groundwork for future research on factuality in LLMs, and position RelationalFactQA as a valuable resource for tracking progress on this critical capability.


\setlength{\tabcolsep}{.6em}
\begin{table}[htbp] 
  \centering
  \small
  \caption{Closed-book QA datasets characteristics\tablefootnote{Obtained from the train set for: WikiSQL, WikiTableQuestions, Open-WikiTable, TAT-QA and NQ-Open.}. Prior datasets have outputs with approximately one tuple and one attribute. In contrast, RelationalFactQA demands  complex outputs, with an average of 27 tuples and 5.3 attributes per answer.}
  \label{tab:data_details}
  \begin{tabular}{lrrrr} 
    \textbf{} & \textbf{Total \#} & \textbf{Avg \#} & \textbf{Avg \#} & \textbf{Avg \#} \\
    \textbf{Dataset} & \textbf{Questions} & \textbf{Output Tuples} & \textbf{Output Attributes} & \textbf{Output Tokens} \\
    \hline
    WikiSQL~\cite{Zhong2017Seq2SQL} & 56,355 & 1.08 & 1.00 & 3.22 \\
    WikiTableQuestions~\cite{Pasupat2015WikiTableQuestions} & 14,149 & 1.08 & 1.00 & 2.80 \\
    Open-WikiTable~\cite{kweon2023openwikitabledatasetopendomain} & 53,819 & 1.08 & 1.00 & 3.23 \\
    TAT-QA~\cite{zhu2021tat} & 13,215 & 1.19 & n.a. & 6.63 \\
    TruthfulQA~\cite{lin2022truthfulqameasuringmodelsmimic} & 790 & 1.00 & n.a.  & 10.49 \\
    TriviaQA (unfiltered)~\cite{joshi2017triviaqalargescaledistantly} & 87,622 & 1.00 & n.a. & 6.39 \\
    NQ-Open~\cite{47761, lee-etal-2019-latent} & 87,925 & 1.22 & n.a. & 4.30 \\
    SimpleQA~\cite{simpleQAOpenAI} & 4,326 & 1.00 & n.a. & 4.20 \\
    \hline
    \textbf{RFQA} & 696 & \textbf{26.942} & \textbf{5.32} & \textbf{357.09} \\
    \hline
  \end{tabular}
\end{table}

\section{Related Work}
\label{sec:related}
The evaluation of factual accuracy in LLMs has lead to the development of diverse benchmarks~\cite{10.1145/3641289}. However, existing work evaluates an LLM's ability to return short-span answers, rather than complex, structured relational data.
As motivated in Section 1, the ability to generate such larger, structured outputs presents distinct challenges beyond single-fact recall, involving sustained coherence and factual consistency across multiple data points \cite{liu-etal-2024-lost, Holtzman2020TheCurious}.
Table~\ref{tab:data_details} provides a comparative overview of output characteristics across several closed-book QA datasets and RelationalFactQA.


\paragraph{Factuality Evaluation Benchmarks.}
A significant body of work focuses on evaluating the factual correctness of LLM generations. Benchmarks such as TriviaQA, NQ-Open, and TruthfulQA assess LLMs' ability to answer questions with short, often single-entity or single-value, factual statements~\cite{simpleQAOpenAI}. While these are crucial for gauging general world knowledge, they do not probe the model's capacity to synthesize answers as structured relations. As evident in Table~\ref{tab:data_details}, the expected outputs in these datasets typically consist of a single tuple and a single attribute. Other efforts like FactScore or HaluEval variants aim to quantify hallucination rates~\cite{Ji2023SurveyHallucination}, but again, within the context of single-statement claims rather than structured relational outputs. 
Despite these varied evaluation efforts, the fundamental challenge of LLM hallucination persists as a critical concern~\cite{chowdhury2025truthfulness,hallReport}.

\paragraph{Table Question Answering and Reasoning.}
Several benchmarks like WikiSQL~\cite{Zhong2017Seq2SQL}, WikiTableQuestions~\cite{Pasupat2015WikiTableQuestions}, and TabFact~\cite{Chen2020TabFact} involve tabular data. However, these benchmarks \textit{provide the relevant table(s) as input} to the LLM, tasking it with understanding, reasoning over, or extracting information from the provided context~\cite{tableBench2025}. 
In contrast, RelationalFactQA operates in a closed-book setting, where the LLM retrieves the tabular answer from its parametric knowledge. 
This shifts the evaluation from context-based reasoning to parametric relational knowledge retrieval.

\paragraph{Text-to-SQL.} 
While RelationalFactQA uses SQL as one input modality to query the LLM's knowledge, our focus is not on the correctness of SQL generation itself, which is the primary goal of Text-to-SQL benchmarks~\cite{yu2018spider,li2024bird,Survey-Text2SQL-LLMs,hong2024survey_sql,saparina2024ambrosia}. Instead, we evaluate the factual accuracy and completeness of the \textit{tabular data returned by the LLM} in response to a query (be it in natural language or SQL). We manually filter examples from two Text2SQL datasets and adapt them to the Relational Fact Retrieval task in building our benchmark. 

\paragraph{Knowledge Probes for LLMs.}
Prior research has explored using ``knowledge probes" (e.g., LAMA~\cite{petroni2019language}) to assess what factual information is stored in an LLM's parameters, typically by prompting models to fill in missing tokens in cloze-style statements (e.g., ``Paris is the capital of [MASK]"). These probes generally target single, atomic facts~\cite{petroni2020how}. RelationalFactQA extends this concept from single-fact elicitation to probing for multi-tuple, multi-attribute relational structures. 


In summary, while existing benchmarks address various facets of LLM factuality, RelationalFactQA fills a critical gap by specifically evaluating LLMs' ability to act as ``parametric databases," retrieving factual information - in contrast with plausible data \cite{borisov2023language} - in a tabular format. 


\section{The Benchmark}
\label{sec:benchmark}

{\bf Task Definition and Problem Formulation.}
We define the task of \textit{Relational Fact Retrieval} as the generation of structured, multi-record, multi-attribute tabular data by an LLM in response to a query, relying exclusively on the model's internal parametric knowledge. 

Formally, the problem is formulated as follows:
\begin{itemize}[leftmargin=*]
    \item \textbf{Input:} The input is a query $q$, which can be expressed either in natural language (NL) or as a Structured Query Language (SQL) statement. The query $q$ specifies the factual information to retrieve and the desired output relational structure.

    \item \textbf{Output:} The desired output is a table $\hat{T}$. This table is characterized by a schema $S = \{A_1, A_2, \dots, A_k\}$, representing $k$ attributes (columns), and a set of $n$ tuples (rows), where $n \ge 0$. Each tuple $t_i \in \hat{T}$ is an ordered list of $k$ cell values $(v_{i1}, v_{i2}, \dots, v_{ik})$, corresponding to the attributes in $S$.
\end{itemize}

 LLMs are instructed to enforce a closed-book evaluation setting and, where applicable, technically restricted, e.g., by disabling access to external tools, web browsing functionalities, or code execution environments via API parameters. The closed-book setting is intentional: in retrieval-augmented generation (RAG) or tool-assisted workflows, the factual quality of outputs depends not only on the model’s internal knowledge, but also on external factors—such as retrieval accuracy, context formatting, or prompt design. These confounding variables make it difficult to isolate the LLM’s intrinsic factual competence. 
While retrieval-based methods may improve factual coverage, we hypothesize the challenges in closed-book persist also in open-book scenarios



\textbf{Dataset Construction.}
To build the RelationalFactQA dataset, we 
combine manual curation and semi-automatic generation. 

In the manual pipeline, we consider 44 datasets from three existing corpora of examples (Spider~\cite{yu2018spider}, Bird~\cite{li2024bird}, and Galois~\cite{galois1}) that contain natural language (NL) and SQL query pairs along with their underlying structured databases. We manually review each dataset in two steps. First, we identify the databases with schema and entities that are present on Wikipedia - this is important to ensure that the examples are within the knowledge scope of an LLM\footnote{While entities have different popularity online, we experimentally verified that this dimension does not impact our experiments.}. 
Second, for each database, we retain only the NL queries that reference factual, world-knowledge content that is temporally stable, deliberately excluding subjective or dynamic information such as user reviews or prices. The corresponding SQL queries and their tabular outputs are finally included in the benchmark.

For the semi-automatic pipeline, we adopt a two-step process: first, we generate tables to serve as query targets; then, we construct corresponding NL question–SQL query pairs. To ensure that the table schemas and entities are likely to be known by LLMs, we extract data from the YAGO 4.5 knowledge base~\cite{suchanek2024yago45largeclean}--a structured resource derived from Wikidata.
YAGO is organized around RDF triplets; each has a subject connected to an object through a predicate, e.g., ``Trump, president, USA" or ``NYC, population, 8.2M". To obtain tables, we follow the procedure of selecting seven Yago types (high level classes, such as City and Country) and reorganizing the triples to collect multiple attributes for those (such as size in squared Km)~\cite{abs-2402-06282}.  

Using an automatic generator tool, Qatch~\cite{papicchio2023qatch}, we then create the corresponding NL-SQL pairs for these YAGO-derived tables. 
To ensure controlled complexity for this segment of the benchmark, the Qatch generation strategy deliberately focuses on \texttt{SELECT} queries. These queries are designed to systematically vary in two main dimensions: the number of projected attributes (columns) and the complexity of selection, achieved by altering the number and nature of predicates in the \texttt{WHERE} clause. Therefore, the Qatch-generated queries predominantly feature projection and filtering operations, allowing for a targeted assessment of these core capabilities.

While the full RelationalFactQA benchmark incorporates a wider range of SQL operators, including \texttt{JOIN} and \texttt{AGGREGATE} functions, these more complex operators are sourced from the manually curated datasets (Spider, Bird, Galois). These human-authored queries contribute crucial linguistic and structural diversity to the benchmark. The rationale for the focused Qatch generation approach, emphasizing projection and selection, is that the primary challenge lies in the LLM’s ability to accurately retrieve the fundamental base data; if this initial extraction is flawed, any subsequent, more complex operations (such as the joins or aggregations found in other parts of the benchmark) would inherently build upon incorrect information. As the tool occasionally produces syntactically correct but semantically trivial or invalid queries, we manually remove such non-meaningful examples.


Finally, we perform targeted preprocessing steps to enhance consistency in the ground truth data. For all date attributes, we extract the year component to ensure that any condition involving dates can be treated as numerical comparisons, rather than requiring models to process full date-type values. Also, we manually removed noisy tuples, such as instances where organizations were listed as Nobel Prize laureates instead of individuals. These actions ensure comparable outputs across samples, focusing the evaluation on the fact retrieval capabilities.


\begin{figure}[htbp]
\centering
\begin{minipage}[t]{0.39\textwidth}
\caption*{(a) Source Distribution}
\centering
\resizebox{0.95\textwidth}{!}{%

\begin{tikzpicture}
\pie[
    text=legend,
    radius=2.0,
    color={blue!50, red!50, green!50, orange!50}
]{
    11/BIRD,
    10/GALOIS,
    71/QATCH,
    8/SPIDER
}
\end{tikzpicture}
}
\end{minipage}
\hfill
\begin{minipage}[t]{0.59\textwidth}
\caption*{(b) Query Complexity Distribution}
\centering
\resizebox{0.6\textwidth}{!}{%
\begin{tabular}{lr} 
    \textbf{{Type}} & \textbf{\# Questions} \\
    \hline
    \textsc{SELECT} without \textsc{WHERE} & 10 \\
    \textsc{WHERE} numerical condition & 148 \\
    \textsc{WHERE} categorical condition & 294 \\
    \textsc{WHERE} mixed condition & 294 \\
    \textsc{AGGREGATIVE} & 49 \\
    \textsc{JOIN} & 67 \\
    \textsc{DISTINCT} & 34 \\
    \textsc{GROUP BY} & 13 \\
    \textsc{LIMIT} & 11\\
    \textsc{ORDER BY} & 17 \\
    \hline
\end{tabular}
}
\end{minipage}
\caption{\textbf{RFQA} dataset. Source distribution and distribution of query complexity (SQL operators).}
\label{fig:rfqa_stats}
\end{figure}

\textbf{Dataset Statistics.}
The RFQA benchmark comprises 696 question, query, answer triples. As reported in Figure~\ref{fig:rfqa_stats}(a), the majority of questions (71\%) are from the Qatch pipeline, ensuring controlled complexity and coverage, while contributions from Bird (11\%), Galois (10\%), and Spider (8\%) provide diverse, human-authored queries. This hybrid approach allows RFQA to cover a range of factual domains, including 
common entities typically found within an LLM's pre-training corpus. 

A key characteristic of RFQA is the size of its target outputs, designed to test an LLM's ability to generate structured relational data. As detailed in Table~\ref{tab:data_details}, ground truth answers in RFQA contain an average of \textbf{357 tokens}, specifically \textbf{26.94 tuples} (rows) and \textbf{5.32 attributes} (columns), for an average of \textbf{135.50 cells} per table. The output dimensions exhibit considerable variability: the number of tuples ranges from a minimum of 1 to a maximum of 904, while attributes span from 1 to 9. This contrasts sharply with prior QA benchmarks, which typically expect single-tuple, single-attribute answers. The attribute types within RFQA tables also vary; on average, each target table schema consists of approximately 1.06 numerical attributes, 3.16 categorical attributes, and 4.26 attributes containing mixed (numerical and string) data types. 

The complexity of the retrieval task is also defined by the SQL constructs associated with each question. Figure~\ref{fig:rfqa_stats}(b) presents the distribution of SQL operators within RFQA. 
The distribution reflects our focus on evaluating the retrieval of 
data under diverse projection and filtering requirements. 

\section{Experimental Settings}

{\bf Retrieval Methods.} We evaluate LLMs on RFQA using three iterative methods:
\begin{itemize}[leftmargin=*]
    \item \textsc{NL}. The LLM is directly prompted with a natural language query $q$, requesting the model to return tabular results based on its internal knowledge.
    \item \textsc{SQL}. Similar to the NL approach, but the query $q$ is expressed in SQL. The model is expected to interpret the SQL semantics and return the corresponding tabular data.
    \item \textsc{CoT}. Given an SQL query $q$, a Chain-of-Thought approach~\cite{galois1} decomposes the query execution into two steps: (1) the LLM is prompted to retrieve the relevant base data (i.e., a broader result set), and (2) relational algebra operations are applied in memory on the intermediate output to produce the final filtered result. This method aims to improve retrieval accuracy by breaking queries into simpler tasks.
\end{itemize}

In all methods, the LLM is prompted with the query $q$ and the corresponding output schema $s$, expressed in JSON Schema format.

{\bf Output Processing.} The prompt includes instructions for the model to return results in valid JSON. If no answer is found, the model is instructed to return an empty JSON object.
Each strategy is applied iteratively. After the initial prompt, if the model returns a non-empty result, it is prompted again to return additional data until the model returns an empty JSON. 
Prompt templates used in the experiments are detailed in the Appendix.

Since LLMs do not always produce outputs in valid JSON format, we apply heuristics to extract and recover structured responses. Our approach begins by identifying all text enclosed between ``\{'' and ``\}'' or ``['' and ``]''. If this content forms a valid JSON object, we parse it directly and return it in a result.
If the content is invalid, we re-prompt for correct formatting or attempt to repair common issues like syntax errors or truncation. If recovery fails, the response is treated as invalid. Further details on recovery strategies are in the Appendix.

{\bf Models.} We use open-source and proprietary LLMs. To enhance reproducibility and obtain deterministic results, we set the temperature to 0.0. 
For open-source models, we adopt the following models hosted on Together.AI: Mistral-7B~\cite{jiang2023mistral}, Qwen2.5 and Qwen3~\cite{qiao2024qwen2}, LLama 3 (covering versions 3.1 and 3.3)~\cite{meta2024LLama3}, Gemma 2~\cite{gemma2024gemma}, DeepSeek-LLama3 (as a base for the reasoning model DeepSeek R1 Distill LLama)~\cite{deepseek2024deepseek}. As proprietary models, we use GPT-4.1 and GPT 4.1 mini~\cite{openai2023gpt4}.

{\bf Metrics.}
To evaluate the factuality of each LLM we measure the quality of the produced responses. Each example in the RFQA dataset consists of a query $q$ (either NL or SQL) 
and the corresponding expected set of tuples $t_{exp}$ (the ground-truth).  
To evaluate an LLM, we execute the query $q$ on it and collect the resulting set of tuples $t_{act}$. To assess the quality of the result, we compare tuple sets $t_{exp}$ and $t_{act}$. We adopt two metrics commonly used to benchmark queries executed by LLMs~\cite{papicchio2023qatch}:
\begin{itemize}[leftmargin=*]
\item \textsc{F1}: We compute the F1 score over the set of cells in $t_{act}$ with respect to those in $t_{exp}$. This metric evaluates performance at the cell level, disregarding tuple structure and focusing purely on the correctness of returned values.
\item \textsc{TS} (Tuple Similarity): We measure the fraction of tuples in $t_{exp}$ that also appear in $t_{act}$, comparing tuples holistically. A Tuple Similarity score of 1.0 indicates that $t_{exp}$ and $t_{act}$ share the same schema, cardinality, and cell values. This metric is stricter than \textsc{F1}, as it requires correct grouping of values within tuples, not just correct individual values.
\end{itemize}

To account for superficial differences in formatting (e.g., ``1K'' vs. ``1000''), we normalize all cell values in both $t_{act}$ and $t_{exp}$ before evaluation. This  step mitigates false negatives caused by representational variations. The normalization process involves the following steps: (i) Replacing accented characters with their unaccented equivalents (e.g., ``é'' → ``e''); (ii) Converting all characters to lowercase; 
(iii) Converting shorthand numeric notations like ``1K'' or ``1M'' and into the corresponding numeric values (e.g., ``1K'' → 1000); (iv) Standardizing numeric formats 
(e.g., converting ``1.000,5'' and ``1,000.5'' into a consistent representation).

Moreover, since LLMs may produce answers that are close, but not identical, to the ground truth (e.g., ``Bill Clinton'' vs. ``Bill J. Clinton''), we incorporate approximate matching. Specifically, we use Edit Distance~\cite{ristad1998learning} with a threshold of 10\% relative to the length of the expected string. 
For numerical values, we apply a tolerance of ±10\% relative to the expected number.

To compare two tuples $t_{a}$ and $t_{e}$, we evaluate each pair of corresponding cells based on their shared attribute, using the same comparison strategy as defined previously for the cells. While our current implementation uses simple, efficient matching rules, more advanced approaches such as entity resolution~\cite{papadakis2021four,gadget} or tuple-level instance comparison~\cite{instanceComparisonEDBT} could be applied for more nuanced matching, but
they require manual user configuration and thus cannot be easily used as a metric.



%

\section{Results}
\label{sec:exp}
We organize our evaluation around three main research questions. 
\begin{enumerate}[leftmargin=*]
    \item \textit{Factuality}. To what extent can LLMs generate factual tables based on their internal knowledge?
    \item \textit{Extraction Techniques}. Are LLMs more effective at generating tabular responses from SQL queries compared to NL questions? Does CoT help in getting better results?
    \item \textit{Query complexity}. Do LLMs' performance depend on the schema and the query complexity? 
\end{enumerate}

\newcommand{\colorcellSimple}[1]{%
    \cellcolor{green!\fpeval{#1 * 100}}#1
}

\newcommand{\colorcell}[1]{%
  \begingroup
  \edef\tempval{\fpeval{round(#1, 3)}}%
  \ifdim \tempval pt > 0.5pt
    \cellcolor{green!\fpeval{round(#1*50, 3)}}\tempval
  \else
    \ifdim \tempval pt < 0.2pt
        \cellcolor{red!\fpeval{round((1-#1)*50, 3)}}\tempval
    \else
        \cellcolor{orange!\fpeval{round((1-#1)*50, 3)}}\tempval
    \fi
  \fi
  \endgroup
}


\setlength{\tabcolsep}{.45em}

\begin{table}[htbp] 
\small
  \centering
  \caption{Benchmark Results. F1 and Tuple Similarity (TS) measured for all LLMs in our evaluation. AVG is the average between F1 and TS. LLMs ordered by increasing size in terms of parameters.}
  \label{tab:benchmarkResults}
\begin{tabular}{clccccccccc}
\hline
\textbf{} & \textbf{} & 
\shortstack{Mistral \\ 7B} & 
\shortstack{\rule{0pt}{2.2ex} QWEN \\ 2.5-7B} & 
\shortstack{LLama \\ 3.1-8B} & 
\shortstack{GPT \\ 4.1 mini} & 
\shortstack{Gemma \\ 2-9B} & 
\shortstack{LLama \\ 3.3-70B} & 
\shortstack{DeepSeek \\ 70B} & 
\shortstack{GPT \\ 4.1} & 
\shortstack{QWEN \\ 3-235B} \\
\hline
\multirow{3}{*}{\textbf{NL}} 
& F1 & \colorcell{0.440} & \colorcell{0.487} & \colorcell{0.481} & \colorcell{0.537} & \colorcell{0.557} & \colorcell{0.609} & \colorcell{0.606} & \colorcell{0.654} & \colorcell{0.613} \\
& TS & \colorcell{0.076} & \colorcell{0.085} & \colorcell{0.155} & \colorcell{0.115} & \colorcell{0.107} & \colorcell{0.149} & \colorcell{0.150} & \colorcell{0.247} &  \colorcell{0.225} \\
& AVG & \colorcell{0.258} & \colorcell{0.286} & \colorcell{0.318} & \colorcell{0.326} & \colorcell{0.332} & \colorcell{0.379} & \colorcell{0.378} & \colorcell{0.450} & \colorcell{0.419} \\
\hline
\multirow{3}{*}{\textbf{SQL}} 
& F1 & \colorcell{0.346} & \colorcell{0.459} & \colorcell{0.332} & \colorcell{0.417} & \colorcell{0.571} & \colorcell{0.620} & \colorcell{0.600} & \colorcell{0.388} & \colorcell{0.595} \\
& TS & \colorcell{0.042} & \colorcell{0.079} & \colorcell{0.110} & \colorcell{0.055} & \colorcell{0.123} & \colorcell{0.155} & \colorcell{0.142} & \colorcell{0.096} & \colorcell{0.185} \\
& AVG & \colorcell{0.194} & \colorcell{0.269} & \colorcell{0.221} & \colorcell{0.236} & \colorcell{0.347} & \colorcell{0.387} & \colorcell{0.371} & \colorcell{0.302} & \colorcell{0.390} \\
\hline
\multirow{3}{*}{\textbf{CoT}} 
& F1 & \colorcell{0.477} & \colorcell{0.503} & \colorcell{0.585} & \colorcell{0.638} & \colorcell{0.594} & \colorcell{0.677} & \colorcell{0.646} & \colorcell{0.693} & \colorcell{0.651} \\
& TS & \colorcell{0.090} & \colorcell{0.091} & \colorcell{0.127} & \colorcell{0.120} & \colorcell{0.106} & \colorcell{0.157} & \colorcell{0.168} & \colorcell{0.174} & \colorcell{0.228} \\
& AVG & \colorcell{0.284} & \colorcell{0.297} & \colorcell{0.356} & \colorcell{0.379} & \colorcell{0.350} & \colorcell{0.417} & \colorcell{0.407} & \colorcell{0.433} & \colorcell{0.439} \\
\hline
\end{tabular}
\end{table}

\stitle{Exp-1. Overall Performance}
We evaluate all LLMs in our benchmark using the RFQA dataset and report their performance using the two quality metrics: \textsc{F1} and \textsc{TS}. To provide a single, comparable measure of factual accuracy across models, we also compute the average of \textsc{F1} and \textsc{TS}. 

The results in Table~\ref{tab:benchmarkResults} reveal that increasing the number of model parameters generally leads to improved quality performance (and thus factuality) across all retrieval methods (\textsc{NL}, \textsc{SQL}, and \textsc{CoT}). However, the task remains inherently difficult. While larger models, such as Qwen 3, achieve \textsc{F1} scores above 0.6, this improvement does not translate to accurate tuple-level results. The best \textsc{TS} score is only 0.247, obtained by GPT 4.1, highlighting that even frontier models often return wrong values in output tuples.

This experiment also shows that querying using \textsc{NL} has an edge over \textsc{SQL} in all models, while the \textsc{CoT} approach leads to improved retrieval with all LLMs except GTP 4.1. 

\noindent\fbox{%
    \parbox{\linewidth}{\textit{Takeaways for questions (1) and (2)}:  LLMs still struggle to consistently retrieve structured factual knowledge as complete output tuples. NL outperfoms slightly SQL as a retrieval method, while CoT provides benefits in most settings.  
    }}


\stitle{Exp-2. Performance by Attribute Type.}
To investigate the third research question, we exploit the metadata used to annotate each query $q$ in RFQA. 
In this experiment, we analyze model performance based on the type of attributes in the query output. We divide the queries into two categories: those that return only numerical values and those that return only categorical values. We use the average of the \textsc{F1} and \textsc{TS} scores as the metric.

\newcommand{\colorValue}[1]{%
  \ifnum #1 > 0
    \cellcolor{green!\fpeval{min(100, round(#1))}}#1
  \else
    \cellcolor{red!\fpeval{min(100, round(abs(#1)))}}#1
  \fi
}

\setlength{\tabcolsep}{.45em}

\begin{table}[htbp] 
  \centering
  \small
  \caption{Quality measured as the AVG between F1 and TS w.r.t. type of output attributes.}
  \label{tab:breakdownAttrType}
\begin{tabular}{clccccccccc}
\hline
\textbf{} & \textbf{} & 
\shortstack{Mistral \\ 7B} & 
\shortstack{\rule{0pt}{2.2ex} QWEN \\ 2.5-7B} & 
\shortstack{LLama \\ 3.1-8B} & 
\shortstack{GPT \\ 4.1 mini} & 
\shortstack{Gemma \\ 2-9B} & 
\shortstack{LLama \\ 3.3-70B} & 
\shortstack{DeepSeek \\ 70B} & 
\shortstack{GPT \\ 4.1} & 
\shortstack{QWEN \\ 3-235B} \\
\hline
\multirow{3}{*}{\textbf{NL}} 
& \textbf{Num} & 0.120 & 0.170 & 0.134 & 0.167 & 0.225 & 0.225 & 0.215 & 0.511 & 0.393 \\
& \textbf{Cat} & 0.211 & 0.236 & 0.301 & 0.339 & 0.301 & 0.391 & 0.370 & 0.515  & 0.397\\
& Diff \%      & +\colorValue{76}\%    & +\colorValue{39}\%    & +\colorValue{125}\%    & +\colorValue{103}\%    & +\colorValue{34}\%   & +\colorValue{74}\%    & +\colorValue{72}\%    & +\colorValue{1}\% & +\colorValue{1}\%\\
\hline
\multirow{3}{*}{\textbf{SQL}} 
& \textbf{Num} & 0.065 & 0.178 & 0.164 & 0.250 & 0.255 & 0.267 & 0.276 & 0.302 & 0.410 \\
& \textbf{Cat} & 0.107 & 0.212 & 0.249 & 0.182 & 0.297 & 0.416 & 0.397 & 0.262 & 0.345 \\
& Diff \%      & +\colorValue{65}\%    & +\colorValue{19}\%    & +\colorValue{52}\%   & \colorValue{-27}\%    & +\colorValue{16}\%   & +\colorValue{56}\%    & +\colorValue{44}\%    & \colorValue{-15}\% & \colorValue{-16}\% \\
\hline
\multirow{3}{*}{\textbf{CoT}} 
& \textbf{Num} & 0.263 & 0.239 & 0.222 & 0.333 & 0.340 & 0.331 & 0.402 & 0.530 & 0.439\\
& \textbf{Cat} & 0.254 & 0.262 & 0.307 & 0.395 & 0.293 & 0.432 & 0.398 & 0.464 & 0.383\\
& Diff \%      & \colorValue{-3}\%    & +\colorValue{10}\%    & +\colorValue{38}\%    & +\colorValue{19}\%    & \colorValue{-14}\%   & +\colorValue{31}\%    & \colorValue{-1}\%    & \colorValue{-14} \%  & \colorValue{-13} \%\\
\hline
\end{tabular}
\end{table}

Results in Table~\ref{tab:breakdownAttrType} 
show that extracting categorical values is generally easier for small and medium LLMs than retrieving numerical ones. However, larger models perform better on numerical queries than on categorical ones when using SQL and CoT.

\stitle{Exp-3. Performance by Output Size.}
We focus on the top-3 performing LLMs and analyze how their performance varies with the size of the expected output. We group the results according to: $a)$ the number of attributes requested in the query, and $b)$ the overall output size, measured as the number of expected cells (\#rows $\times$ \#attributes). 
We use the \textsc{TS} metric, which accounts for both the structure and completeness of the returned data.

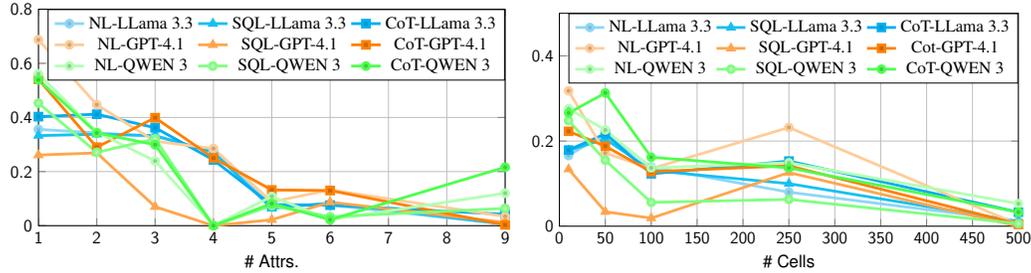
\begin{figure}[ht]
    \begin{minipage}[t]{0.49\textwidth}
        \centering
        \resizebox{1.0\textwidth}{!}{%
\begin{tikzpicture}
  \begin{axis}[
    xlabel={\# Attrs.},
    grid=major,
    ymin=0.0, ymax=0.8,
    xmin=1.0, xmax=9.0,
    legend style = {at={(0.02,1.0)}, legend columns=3, anchor=north west, fill=white,align=left},
    width=11.1cm,
    height=6.0cm,
    cycle multiindex* list={
      mark list*\nextlist
      color list
    }
  ]

    \addplot+[color=cyan!40, mark=* ,line width=1.64pt] coordinates {(1, 0.356) (2,0.343) (3,0.330) (4,0.272) (5,0.080) (6,0.076) (9,0.006)};
    \addlegendentry{NL-LLama 3.3}

    \addplot+[color=cyan!70, mark=triangle* ,line width=1.64pt] coordinates {(1, 0.333) (2,0.338) (3,0.333) (4,0.272) (5,0.064) (6,0.084) (9,0.006)};
    \addlegendentry{SQL-LLama 3.3}

    \addplot+[color=cyan!100, mark=square* ,line width=1.64pt] coordinates {(1, 0.403) (2,0.412) (3,0.362) (4,0.243) (5,0.076) (6,0.076) (9,0.042)};
    \addlegendentry{CoT-LLama 3.3}

    \addplot+[color=orange!40, mark=* ,line width=1.64pt] coordinates {(1, 0.687) (2, 0.447) (3, 0.312) (4, 0.285) (5, 0.088) (6, 0.132) (9, 0.034)};
    \addlegendentry{NL-GPT-4.1}

    \addplot+[color=orange!70, mark=triangle* ,line width=1.64pt] coordinates {(1, 0.261) (2, 0.269) (3, 0.07) (4, 0.0) (5, 0.022) (6, 0.087) (9, 0.012)}; 
    \addlegendentry{SQL-GPT-4.1}

    \addplot+[color=orange!100, mark=square* ,line width=1.64pt] coordinates {(1, 0.541) (2, 0.29) (3, 0.399) (4, 0.25) (5, 0.132) (6, 0.13) (9, 0.003)}; 
    \addlegendentry{CoT-GPT-4.1}

    \addplot+[color=green!30, mark=* ,line width=1.64pt] coordinates {(1, 0.561) (2, 0.346) (3, 0.239) (4, 0.0) (5, 0.108) (6, 0.022) (9, 0.12)};
    \addlegendentry{NL-QWEN 3}
    
    \addplot+[color=green!50, mark=* ,line width=1.64pt] coordinates {(1, 0.453) (2, 0.271) (3, 0.321) (4, 0.0) (5, 0.074) (6, 0.033) (9, 0.064)};
    \addlegendentry{SQL-QWEN 3}

    \addplot+[color=green!70, mark=* ,line width=1.64pt] coordinates {(1, 0.539) (2, 0.343) (3, 0.299) (4, 0.0) (5, 0.082) (6, 0.022) (9, 0.216)};
    \addlegendentry{CoT-QWEN 3}

  \end{axis}
\end{tikzpicture}
        }
    \end{minipage}
    \begin{minipage}[t]{0.49\textwidth}
        \centering
        \resizebox{1.0\textwidth}{!}{%
\begin{tikzpicture}
  \begin{axis}[
    xlabel={\# Cells},
    grid=major,
    ymin=0.0, ymax=0.5,
    xmin=0, xmax=500,
    legend style = {at={(0.02,1.0)}, legend columns=3, anchor=north west, fill=white,align=left},
    width=11.1cm,
    height=6.0cm,
    cycle multiindex* list={
      mark list*\nextlist
      color list
    }
  ]

    \addplot+[color=cyan!40, mark=*,line width=1.64pt] coordinates {(10, 0.166) (50, 0.218) (100, 0.139) (250, 0.08) (500, 0.012)};
    \addlegendentry{NL-LLama 3.3}

    \addplot+[color=cyan!70, mark=triangle*,line width=1.64pt] coordinates {(10, 0.178) (50, 0.205) (100, 0.131) (250, 0.1)  (500, 0.011)};
    \addlegendentry{SQL-LLama 3.3}

    \addplot+[color=cyan!100, mark=square* ,line width=1.64pt] coordinates {(10, 0.179) (50, 0.217) (100, 0.123) (250, 0.153) (500, 0.033)};
    \addlegendentry{CoT-LLama 3.3}

    \addplot+[color=orange!40, mark=*,line width=1.64pt] coordinates {((10, 0.318) (50, 0.172) (100, 0.135) (250, 0.232) (500, 0.004)};
    \addlegendentry{NL-GPT-4.1}

    \addplot+[color=orange!70, mark=triangle* ,line width=1.64pt] coordinates {((10, 0.134) (50, 0.034) (100, 0.019) (250, 0.126) (500, 0.000)};
    \addlegendentry{SQL-GPT-4.1}

    \addplot+[color=orange!100, mark=square* ,line width=1.64pt] coordinates {(10, 0.223) (50, 0.188) (100, 0.128) (250, 0.142) (500, 0.005)};
    \addlegendentry{Cot-GPT-4.1}

    \addplot+[color=green!30, mark=* ,line width=1.64pt] coordinates {(10, 0.276) (50, 0.226) (100, 0.137) (250, 0.149) (500, 0.053)};
    \addlegendentry{NL-QWEN 3}
    
    \addplot+[color=green!50, mark=* ,line width=1.64pt] coordinates {(10, 0.248) (50, 0.155) (100, 0.056) (250, 0.063) (500, 0.006)};
    \addlegendentry{SQL-QWEN 3}

    \addplot+[color=green!70, mark=* ,line width=1.64pt] coordinates {(10, 0.266) (50, 0.313) (100, 0.162) (250, 0.137) (500, 0.033)};
    \addlegendentry{CoT-QWEN 3}

  \end{axis}
\end{tikzpicture}
        }
    \end{minipage}
\caption{TS results for LLama 3.3, GPT-4.1 and QWEN 3, with all retrieval techniques, w.r.t. the expected output measure as the number of attributes (left) and cells (right).}
\label{fig:attributes}
\vspace{-1ex}
\end{figure}

Figure~\ref{fig:attributes} summarizes our findings. On the left side, we show how quality decreases as the number of requested attributes increases, indicating that LLMs struggle more when asked to retrieve wider tables. On the right side, we plot the \textsc{TS} score against the total number of expected cells. The trend remains consistent: as the number of rows and columns grows, the model's ability to return accurate, complete tabular data declines.

\setlength{\tabcolsep}{.45em}
\begin{table}[htbp] 
  \centering
  \small
  \caption{Quality measured as the AVG between F1 and TS w.r.t. query complexity.}
  \label{tab:breakdownQueryComplexity}

\begin{tabular}{l|ccc|ccc|ccc} 
    & \multicolumn{3}{c}{\textbf{LLama 3.3-70B}} & \multicolumn{3}{c}{\textbf{GPT 4.1}}  & \multicolumn{3}{c}{\textbf{QWEN 3-235B}}\\
    \textbf{{Type}} & \textbf{NL} & \textbf{SQL}  & \textbf{CoT} & \textbf{NL} & \textbf{SQL}  & \textbf{CoT} & \textbf{NL} & \textbf{SQL}  & \textbf{CoT} \\
    \hline
 \textsc{SELECT} without \textsc{WHERE} 		& \colorcell{0.599}	&	\colorcell{0.646}	&	\colorcell{0.595} & \colorcell{0.845} & \colorcell{0.399} & \colorcell{0.995} & \colorcell{0.787} & \colorcell{0.582} & \colorcell{0.694} \\
 \textsc{WHERE} numerical condition 			& \colorcell{0.365}	&	\colorcell{0.360}	&	\colorcell{0.387} & \colorcell{0.390} & \colorcell{0.197} & \colorcell{0.382} & \colorcell{0.420} & \colorcell{0.384} & \colorcell{0.444} \\
 \textsc{WHERE} categorical condition 			& \colorcell{0.383}	&	\colorcell{0.393}	&	\colorcell{0.437} & \colorcell{0.466} & \colorcell{0.252} & \colorcell{0.453} & \colorcell{0.414} & \colorcell{0.397} & \colorcell{0.444} \\
 \textsc{WHERE} mixed condition 				& \colorcell{0.376}	&	\colorcell{0.384}	&	\colorcell{0.414} & \colorcell{0.444} & \colorcell{0.240} & \colorcell{0.424} & \colorcell{0.414} & \colorcell{0.387} & \colorcell{0.435} \\
 \textsc{AGGREGATIVE} 							& \colorcell{0.237}	&	\colorcell{0.260}	&	\colorcell{0.254} & \colorcell{0.447} & \colorcell{0.229} & \colorcell{0.353} & \colorcell{0.366} & \colorcell{0.364} & \colorcell{0.259} \\
 \textsc{JOIN} 									& \colorcell{0.143}	&	\colorcell{0.184}	&	\colorcell{0.157} & \colorcell{0.461} & \colorcell{0.126} & \colorcell{0.091} & \colorcell{0.356} & \colorcell{0.222} & \colorcell{0.139} \\
 \textsc{DISTINCT} 								& \colorcell{0.344}	&	\colorcell{0.429}	&	\colorcell{0.354} & \colorcell{0.536} & \colorcell{0.172} & \colorcell{0.319} & \colorcell{0.446} & \colorcell{0.247} & \colorcell{0.342} \\
 \textsc{GROUP BY} 								& \colorcell{0.234}	&	\colorcell{0.228}	&	\colorcell{0.461} & \colorcell{0.325} & \colorcell{0.166} & \colorcell{0.260} & \colorcell{0.181} & \colorcell{0.266} & \colorcell{0.239} \\
 \textsc{LIMIT} 								& \colorcell{0.108}	&	\colorcell{0.189}	&	\colorcell{0.517} & \colorcell{0.417} & \colorcell{0.399} & \colorcell{0.355} & \colorcell{0.146} & \colorcell{0.162} & \colorcell{0.206} \\
 \textsc{ORDER BY} 								& \colorcell{0.174}	&	\colorcell{0.206}	&	\colorcell{0.424} & \colorcell{0.441} & \colorcell{0.349} & \colorcell{0.437} & \colorcell{0.222} & \colorcell{0.179} & \colorcell{0.376} \\
    \hline
  \end{tabular}
\end{table}

\stitle{Exp-4. Query Complexity.}
Table~\ref{tab:breakdownQueryComplexity} provides a breakdown of performance w.r.t. the query complexity. We observe that as query complexity increases, the quality of the generated responses tends to decrease. Simple queries such as \textsc{SELECT} without \textsc{WHERE} consistently achieve the highest scores, while complex constructs like \textsc{JOIN}, \textsc{AGGREGATE}, or multi-condition \textsc{WHERE} clauses report substantially lower results across all models and retrieval methods.
In particular, the \textsc{JOIN} operator represents a notable challenge. Scores are low for all models, especially in the \textsc{CoT} setting as 
it does not yet support joins over multiple tables. 
Despite its limitations, the \textsc{CoT} strategy demonstrates meaningful gains for several complex operations. 
This highlights the benefit of breaking down query execution into intermediate reasoning steps.
Finally, certain operators such as \textsc{LIMIT} and \textsc{ORDER BY} appear systematically difficult for all models and prompting strategies. These constructs require precise handling of position and ordering in the output tuples — capabilities that autoregressive models struggle to maintain as the result set grows.

\noindent\fbox{%
    \parbox{0.98\linewidth}{\textit{Takeaway for question (3)}: LLM performance is significantly influenced by the structure of the target schema. Both attribute type and output size are key factors in determining LLM effectiveness for tabular factual retrieval. Query complexity has also a significant impact on LLM performance.} 
}

\stitle{Results Discussion.}
The challenges observed in generating extensive and accurate tabular data from parametric memory resonate with known LLMs' limitations in long-sequence generation. While issues such as maintaining thematic coherence \cite{liu-etal-2024-lost}, mitigating factual drift \cite{Ji2023SurveyHallucination}, and managing error propagation in autoregressive systems \cite{Holtzman2020TheCurious} are recognized in tasks involving lengthy free-form text, the generation of tabular outputs magnifies these problems. Specifically, the dual axes of table ``size'' - the number of rows (tuples) and the number of columns (attributes) - impose distinct pressures on the model's generative capabilities.

Our findings suggest that the demand for concurrent retrieval and precise alignment of numerous facts strains the model's effective ``working memory'' or its ability to maintain sustained attention to all constraints of the query \cite{liu-etal-2024-lost}. The fact that LLMs often correctly retrieve individual facts in point-wise queries (e.g., the area of a specific county that is reported incorrectly in a larger table) underscores that the bottleneck is frequently not an absence of the underlying factual knowledge. Instead, the difficulty lies in the process of \textit{composing} the individual pieces of information into a larger relational structure. This distinction points towards limitations in the architectural or learned capabilities for synthesis from parameters, rather than simply gaps in memorized knowledge. The dense factual requirement of tabular data, where each cell represents a correct assertion, and the inflexible nature of its structural integrity, make it a valuable testbed for these aspects of LLM performance, revealing failure modes that are less explicitly quantifiable in unstructured generation tasks~\cite{coli_a_00561}.



\section{Conclusions}
\label{sec:concl}



{RelationalFactQA} fills a gap in the factuality landscape by probing LLMs’ ability to act as \emph{parametric databases}: given only a natural-language question or SQL query, a model must assemble multi-row, multi-attribute tables directly from its internal weights.  
Our experiments — with nine LLMs and three querying
modalities — show three consistent trends:

\begin{itemize}[leftmargin=*]
    \item \textbf{Scale helps, but does not solve the problem.} Even the strongest systems score below 0.25 in tuple accuracy, with quality falling sharply as the requested table widens or lengthens.
    \item \textbf{Structure amplifies failure modes.} Errors that remain latent in point-wise QA become evident when multiple cells must be emitted coherently.
    \item \textbf{Prompting matters.} Chain-of-Thought decomposition improves cell-level recall, yet fails to repair tuple mis-alignment and JOIN reasoning.
\end{itemize}

These results underscore that current LLMs retain abundant factual fragments, but lack the mechanisms to \emph{reliably} compose them into relational form.  We release the data, evaluation suite, and prompting templates to support research on (i) architecture changes that improve structured recall, (ii) inference-time strategies for tuple alignment, and (iii) multilingual, temporal, and bias-aware extensions of the task.  We hope {RelationalFactQA} becomes a key resource for
measuring progress toward LLMs that are not just eloquent, but also factual.

\begin{table}[h]
  \caption{Key limitations of \textsc{RelationalFactQA}.}\centering
  \small
  \begin{tabular}{@{}p{2cm}p{5.5cm}p{6.2cm}@{}}
    \toprule
    \textbf{Aspect} & \textbf{Current scope} & \textbf{Implication} \\
    \midrule
    Temporal coverage & Benchmark built from static snapshots; fast-changing facts intentionally excluded. & Cannot assess models’ ability to reason over time-dependent knowledge (e.g., “current GDP”, “latest mayor”). \\
    Linguistic \& cultural breadth & Tables and questions sourced almost exclusively from English-language, Western-centric resources. & Reported performance may not generalize to other languages or under-represented knowledge domains, introducing geographic-cultural bias. \\
    Evaluation granularity & Cell-level scoring uses exact / approximate string and numeric matching with basic normalization. & Semantically correct but lexically different answers (synonyms, alternative units, name variants) can be penalized, under-estimating true capability. \\
    \bottomrule
  \end{tabular}
  \label{tab:limitations}
\end{table}

\textbf{Limitations.} Table~\ref{tab:limitations} summarizes the principal limitations of the current benchmark and outlines how each one affects  our results.

\bibliographystyle{abbrvnat}
\bibliography{biblio}

\begin{thebibliography}{55}
\providecommand{\natexlab}[1]{#1}
\providecommand{\url}[1]{\texttt{#1}}
\expandafter\ifx\csname urlstyle\endcsname\relax
  \providecommand{\doi}[1]{doi: #1}\else
  \providecommand{\doi}{doi: \begingroup \urlstyle{rm}\Url}\fi

\bibitem[{All About AI}(2025)]{hallReport}
{All About AI}.
\newblock Ai hallucination report 2025, 2025.
\newblock URL \url{https://www.allaboutai.com/resources/ai-statistics/ai-hallucinations/}.

\bibitem[Aly et~al.(2021)Aly, Guo, Schlichtkrull, Thorne, Vlachos, Christodoulopoulos, Cocarascu, and Li]{Aly2021FEVEROUS}
R.~Aly, Z.~Guo, M.~S. Schlichtkrull, J.~Thorne, A.~Vlachos, C.~Christodoulopoulos, O.~Cocarascu, and G.~Li.
\newblock {FEVEROUS}: Fact extraction and {VERification} over unstructured and structured information.
\newblock In \emph{Thirty-fifth Conference on Neural Information Processing Systems Datasets and Benchmarks Track (Round 2)}, 2021.

\bibitem[Balsiger et~al.(2024)Balsiger, Dimmler, Egger-Horstmann, and Hanne]{computers13100257}
D.~Balsiger, H.-R. Dimmler, S.~Egger-Horstmann, and T.~Hanne.
\newblock Assessing large language models used for extracting table information from annual financial reports.
\newblock \emph{Computers}, 13\penalty0 (10), 2024.
\newblock ISSN 2073-431X.
\newblock \doi{10.3390/computers13100257}.
\newblock URL \url{https://www.mdpi.com/2073-431X/13/10/257}.

\bibitem[Bisercic et~al.(2023)Bisercic, Nikolic, van~der Schaar, Delibasic, Lio, and Petrovic]{bisercic2023interpretablemedicaldiagnosticsstructured}
A.~Bisercic, M.~Nikolic, M.~van~der Schaar, B.~Delibasic, P.~Lio, and A.~Petrovic.
\newblock Interpretable medical diagnostics with structured data extraction by large language models, 2023.
\newblock URL \url{https://arxiv.org/abs/2306.05052}.

\bibitem[Borisov et~al.(2023)Borisov, Sessler, Leemann, Pawelczyk, and Kasneci]{borisov2023language}
V.~Borisov, K.~Sessler, T.~Leemann, M.~Pawelczyk, and G.~Kasneci.
\newblock Language models are realistic tabular data generators.
\newblock In \emph{The Eleventh International Conference on Learning Representations}, 2023.
\newblock URL \url{https://openreview.net/forum?id=cEygmQNOeI}.

\bibitem[Buoncristiano et~al.(2024)Buoncristiano, Mecca, Santoro, and Veltri]{gadget}
M.~Buoncristiano, G.~Mecca, D.~Santoro, and E.~Veltri.
\newblock Detective gadget: Generic iterative entity resolution over dirty data.
\newblock \emph{Data}, 2024.
\newblock \doi{10.3390/data9120139}.

\bibitem[Cappuzzo et~al.(2024)Cappuzzo, Varoquaux, Coelho, and Papotti]{abs-2402-06282}
R.~Cappuzzo, G.~Varoquaux, A.~Coelho, and P.~Papotti.
\newblock Retrieve, merge, predict: Augmenting tables with data lakes.
\newblock \emph{CoRR}, abs/2402.06282, 2024.
\newblock \doi{10.48550/ARXIV.2402.06282}.
\newblock URL \url{https://doi.org/10.48550/arXiv.2402.06282}.

\bibitem[Chang et~al.(2024)Chang, Wang, Wang, Wu, Yang, Zhu, Chen, Yi, Wang, Wang, Ye, Zhang, Chang, Yu, Yang, and Xie]{10.1145/3641289}
Y.~Chang, X.~Wang, J.~Wang, Y.~Wu, L.~Yang, K.~Zhu, H.~Chen, X.~Yi, C.~Wang, Y.~Wang, W.~Ye, Y.~Zhang, Y.~Chang, P.~S. Yu, Q.~Yang, and X.~Xie.
\newblock A survey on evaluation of large language models.
\newblock \emph{ACM Trans. Intell. Syst. Technol.}, 15\penalty0 (3), mar 2024.
\newblock ISSN 2157-6904.
\newblock \doi{10.1145/3641289}.
\newblock URL \url{https://doi.org/10.1145/3641289}.

\bibitem[Chen et~al.(2020)Chen, Lilley, Gu, Qian, Zhong, Gimpel, and Toutanova]{Chen2020TabFact}
W.~Chen, A.~Lilley, J.~Gu, Z.~Qian, V.~Zhong, K.~Gimpel, and K.~Toutanova.
\newblock Tabfact: A large-scale dataset for table-based fact verification.
\newblock In \emph{International Conference on Learning Representations (ICLR)}, 2020.

\bibitem[Chowdhury et~al.(2025)Chowdhury, Johnson, Huang, Steinhardt, and Schwettmann]{chowdhury2025truthfulness}
N.~Chowdhury, D.~Johnson, V.~Huang, J.~Steinhardt, and S.~Schwettmann.
\newblock Investigating truthfulness in a pre-release o3 model.
\newblock \url{https://transluce.org/investigating-o3-truthfulness}, April 2025.

\bibitem[Christophides et~al.(2021)Christophides, Efthymiou, Palpanas, Papadakis, and Stefanidis]{papadakis2021four}
V.~Christophides, V.~Efthymiou, T.~Palpanas, G.~Papadakis, and K.~Stefanidis.
\newblock An overview of end-to-end entity resolution for big data.
\newblock \emph{{ACM} Comput. Surv.}, 53\penalty0 (6):\penalty0 127:1--127:42, 2021.

\bibitem[{DeepSeek AI}(2024)]{deepseek2024deepseek}
{DeepSeek AI}.
\newblock Deepseek llm: Scaling open-source language models with longtermism.
\newblock \emph{arXiv preprint arXiv:2401.02954}, 2024.
\newblock \doi{10.48550/ARXIV.2401.02954}.
\newblock URL \url{https://doi.org/10.48550/arXiv.2401.02954}.

\bibitem[Dong et~al.(2024)Dong, Stratopoulos, and Wang]{DONG2024100715}
M.~M. Dong, T.~C. Stratopoulos, and V.~X. Wang.
\newblock A scoping review of chatgpt research in accounting and finance.
\newblock \emph{International Journal of Accounting Information Systems}, 55:\penalty0 100715, 2024.
\newblock ISSN 1467-0895.
\newblock \doi{https://doi.org/10.1016/j.accinf.2024.100715}.
\newblock URL \url{https://www.sciencedirect.com/science/article/pii/S1467089524000484}.

\bibitem[Elnashar et~al.(2025)Elnashar, White, and Schmidt]{frai.2025.1558938}
A.~Elnashar, J.~White, and D.~C. Schmidt.
\newblock Enhancing structured data generation with gpt-4o evaluating prompt efficiency across prompt styles.
\newblock \emph{Frontiers in Artificial Intelligence}, Volume 8 - 2025, 2025.
\newblock ISSN 2624-8212.
\newblock \doi{10.3389/frai.2025.1558938}.
\newblock URL \url{https://www.frontiersin.org/journals/artificial-intelligence/articles/10.3389/frai.2025.1558938}.

\bibitem[Gao et~al.(2025)Gao, Hu, Yin, Ruan, Pu, and Wan]{coli_a_00561}
M.~Gao, X.~Hu, X.~Yin, J.~Ruan, X.~Pu, and X.~Wan.
\newblock Llm-based nlg evaluation: Current status and challenges.
\newblock \emph{Computational Linguistics}, pages 1--28, 04 2025.
\newblock ISSN 0891-2017.
\newblock \doi{10.1162/coli_a_00561}.
\newblock URL \url{https://doi.org/10.1162/coli\_a\_00561}.

\bibitem[{Gemma Team, Google}(2024)]{gemma2024gemma}
{Gemma Team, Google}.
\newblock Gemma: Open models based on gemini research and technology.
\newblock \emph{arXiv preprint arXiv:2403.08295}, 2024.
\newblock \doi{10.48550/ARXIV.2403.08295}.
\newblock URL \url{https://doi.org/10.48550/arXiv.2403.08295}.

\bibitem[Glavic et~al.(2024)Glavic, Mecca, Miller, Papotti, Santoro, and Veltri]{instanceComparisonEDBT}
B.~Glavic, G.~Mecca, R.~J. Miller, P.~Papotti, D.~Santoro, and E.~Veltri.
\newblock Similarity measures for incomplete database instances.
\newblock In L.~Tanca, Q.~Luo, G.~Polese, L.~Caruccio, X.~Oriol, and D.~Firmani, editors, \emph{Proceedings 27th International Conference on Extending Database Technology, {EDBT} 2024, Paestum, Italy, March 25 - March 28}, pages 461--473. OpenProceedings.org, 2024.
\newblock \doi{10.48786/EDBT.2024.40}.
\newblock URL \url{https://doi.org/10.48786/edbt.2024.40}.

\bibitem[Holtzman et~al.(2020)Holtzman, Buys, Du, Forbes, and Choi]{Holtzman2020TheCurious}
A.~Holtzman, J.~Buys, L.~Du, M.~Forbes, and Y.~Choi.
\newblock The curious case of neural text degeneration.
\newblock In \emph{International Conference on Learning Representations}, 2020.
\newblock URL \url{https://openreview.net/forum?id=rygGQyrFvH}.

\bibitem[Hong et~al.(2024)Hong, Yuan, Zhang, Chen, Dong, Huang, and Huang]{hong2024survey_sql}
Z.~Hong, Z.~Yuan, Q.~Zhang, H.~Chen, J.~Dong, F.~Huang, and X.~Huang.
\newblock Next-generation database interfaces: A survey of llm-based text-to-sql.
\newblock \emph{arXiv preprint arXiv:2406.08426}, 2024.

\bibitem[Ji et~al.(2023)Ji, Lee, Frieske, Yu, Su, Xu, Ishii, Bang, Madotto, and Fung]{Ji2023SurveyHallucination}
Z.~Ji, N.~Lee, R.~Frieske, T.~Yu, D.~Su, Y.~Xu, E.~Ishii, Y.~J. Bang, A.~Madotto, and P.~Fung.
\newblock Survey of hallucination in natural language generation.
\newblock \emph{ACM Comput. Surv.}, 55\penalty0 (12), Mar. 2023.
\newblock ISSN 0360-0300.
\newblock \doi{10.1145/3571730}.
\newblock URL \url{https://doi.org/10.1145/3571730}.

\bibitem[Jiang et~al.(2023)Jiang, Sablayrolles, Mensch, Bamford, Chaplot, Casas, Bressand, Lengyel, Lample, Saulnier, et~al.]{jiang2023mistral}
A.~Q. Jiang, A.~Sablayrolles, A.~Mensch, C.~Bamford, D.~S. Chaplot, D.~d.~l. Casas, F.~Bressand, G.~Lengyel, G.~Lample, L.~Saulnier, et~al.
\newblock Mistral 7b.
\newblock \emph{arXiv preprint arXiv:2310.06825}, 2023.
\newblock \doi{10.48550/ARXIV.2310.06825}.
\newblock URL \url{https://doi.org/10.48550/arXiv.2310.06825}.

\bibitem[Joshi et~al.(2017)Joshi, Choi, Weld, and Zettlemoyer]{joshi2017triviaqalargescaledistantly}
M.~Joshi, E.~Choi, D.~S. Weld, and L.~Zettlemoyer.
\newblock Triviaqa: A large scale distantly supervised challenge dataset for reading comprehension, 2017.
\newblock URL \url{https://arxiv.org/abs/1705.03551}.

\bibitem[Kasneci et~al.(2023)Kasneci, Se{\ss}ler, Kuchemann, Bannert, Dementieva, Fischer, Gasser, Groh, Hahnel, Hett, Hett, K{\"a}rger, Liu, Liu, Nerdel, Nistor, Scheid, Stallasch, Stober, and Kasneci]{Kasneci2023ChatGPTEducation}
E.~Kasneci, K.~Se{\ss}ler, S.~Kuchemann, M.~Bannert, D.~Dementieva, F.~Fischer, U.~Gasser, G.~Groh, G.~Hahnel, M.~C. Hett, N.-E. Hett, N.~K{\"a}rger, J.~Liu, X.~Liu, M.~Nerdel, J.~Nistor, C.~Scheid, R.~Stallasch, S.~Stober, and G.~Kasneci.
\newblock {ChatGPT for good? On opportunities and challenges of large language models for education}.
\newblock \emph{Learning and Individual Differences}, 103:\penalty0 102274, 2023.
\newblock This article discusses both the potential benefits and the significant challenges, including accuracy and reliability, of using LLMs like ChatGPT in education.

\bibitem[Kweon et~al.(2023)Kweon, Kwon, Cho, Jo, and Choi]{kweon2023openwikitabledatasetopendomain}
S.~Kweon, Y.~Kwon, S.~Cho, Y.~Jo, and E.~Choi.
\newblock Open-wikitable: Dataset for open domain question answering with complex reasoning over table, 2023.
\newblock URL \url{https://arxiv.org/abs/2305.07288}.

\bibitem[Kwiatkowski et~al.(2019)Kwiatkowski, Palomaki, Redfield, Collins, Parikh, Alberti, Epstein, Polosukhin, Kelcey, Devlin, Lee, Toutanova, Jones, Chang, Dai, Uszkoreit, Le, and Petrov]{47761}
T.~Kwiatkowski, J.~Palomaki, O.~Redfield, M.~Collins, A.~Parikh, C.~Alberti, D.~Epstein, I.~Polosukhin, M.~Kelcey, J.~Devlin, K.~Lee, K.~N. Toutanova, L.~Jones, M.-W. Chang, A.~Dai, J.~Uszkoreit, Q.~Le, and S.~Petrov.
\newblock Natural questions: a benchmark for question answering research.
\newblock \emph{Transactions of the Association of Computational Linguistics}, 2019.

\bibitem[Lee et~al.(2019)Lee, Chang, and Toutanova]{lee-etal-2019-latent}
K.~Lee, M.-W. Chang, and K.~Toutanova.
\newblock Latent retrieval for weakly supervised open domain question answering.
\newblock In A.~Korhonen, D.~Traum, and L.~M{\`a}rquez, editors, \emph{Proceedings of the 57th Annual Meeting of the Association for Computational Linguistics}, pages 6086--6096, Florence, Italy, July 2019. Association for Computational Linguistics.
\newblock \doi{10.18653/v1/P19-1612}.
\newblock URL \url{https://aclanthology.org/P19-1612/}.

\bibitem[Li et~al.(2024)Li, Hui, Qu, Yang, Li, Li, Wang, Qin, Geng, Huo, et~al.]{li2024bird}
J.~Li, B.~Hui, G.~Qu, J.~Yang, B.~Li, B.~Li, B.~Wang, B.~Qin, R.~Geng, N.~Huo, et~al.
\newblock Can llm already serve as a database interface? a big bench for large-scale database grounded text-to-sqls.
\newblock \emph{Advances in Neural Information Processing Systems}, 36, 2024.

\bibitem[Lin et~al.(2022)Lin, Hilton, and Evans]{lin2022truthfulqameasuringmodelsmimic}
S.~Lin, J.~Hilton, and O.~Evans.
\newblock Truthfulqa: Measuring how models mimic human falsehoods, 2022.
\newblock URL \url{https://arxiv.org/abs/2109.07958}.

\bibitem[Liu et~al.(2025)Liu, Russo, Cafarella, Cao, Chen, Chen, Franklin, Kraska, Madden, Shahout, and Vitagliano]{palimpzestCIDR}
C.~Liu, M.~Russo, M.~Cafarella, L.~Cao, P.~B. Chen, Z.~Chen, M.~Franklin, T.~Kraska, S.~Madden, R.~Shahout, and G.~Vitagliano.
\newblock Palimpzest: Optimizing ai-powered analytics with declarative query processing.
\newblock In \emph{Proceedings of the {{Conference}} on {{Innovative Database Research}} ({{CIDR}})}, 2025.

\bibitem[Liu et~al.(2024{\natexlab{a}})Liu, Lin, Hewitt, Paranjape, Bevilacqua, Petroni, and Liang]{liu-etal-2024-lost}
N.~F. Liu, K.~Lin, J.~Hewitt, A.~Paranjape, M.~Bevilacqua, F.~Petroni, and P.~Liang.
\newblock Lost in the middle: How language models use long contexts.
\newblock \emph{Transactions of the Association for Computational Linguistics}, 12:\penalty0 157--173, 2024{\natexlab{a}}.
\newblock \doi{10.1162/tacl_a_00638}.
\newblock URL \url{https://aclanthology.org/2024.tacl-1.9/}.

\bibitem[Liu et~al.(2024{\natexlab{b}})Liu, Shen, Li, Ma, Jiang, Luo, Zhang, Fan, Li, and Tang]{Survey-Text2SQL-LLMs}
X.~Liu, S.~Shen, B.~Li, P.~Ma, R.~Jiang, Y.~Luo, Y.~Zhang, J.~Fan, G.~Li, and N.~Tang.
\newblock A survey of {NL2SQL} with large language models: Where are we, and where are we going?
\newblock \emph{CoRR}, abs/2408.05109, 2024{\natexlab{b}}.
\newblock \doi{10.48550/ARXIV.2408.05109}.
\newblock URL \url{https://doi.org/10.48550/arXiv.2408.05109}.

\bibitem[{Meta AI}(2024)]{meta2024LLama3}
{Meta AI}.
\newblock The llama 3 herd of models, 2024.
\newblock URL \url{https://arxiv.org/abs/2407.21783}.

\bibitem[OpenAI(2023)]{openai2023gpt4}
OpenAI.
\newblock Gpt-4 technical report.
\newblock \emph{arXiv preprint arXiv:2303.08774}, 2023.
\newblock \doi{10.48550/ARXIV.2303.08774}.
\newblock URL \url{https://doi.org/10.48550/arXiv.2303.08774}.

\bibitem[Papicchio et~al.(2023)Papicchio, Papotti, and Cagliero]{papicchio2023qatch}
S.~Papicchio, P.~Papotti, and L.~Cagliero.
\newblock Qatch: Benchmarking sql-centric tasks with table representation learning models on your data.
\newblock \emph{Advances in Neural Information Processing Systems}, 36:\penalty0 30898--30917, 2023.

\bibitem[Pasupat and Liang(2015)]{Pasupat2015WikiTableQuestions}
P.~Pasupat and P.~Liang.
\newblock Compositional semantic parsing on semi-structured tables.
\newblock In \emph{Proceedings of the 53rd Annual Meeting of the Association for Computational Linguistics and the 7th International Joint Conference on Natural Language Processing (Volume 1: Long Papers)}, pages 1470--1480, 2015.

\bibitem[Patel et~al.(2025)Patel, Jha, Pan, Gupta, Asawa, Guestrin, and Zaharia]{lotus}
L.~Patel, S.~Jha, M.~Pan, H.~Gupta, P.~Asawa, C.~Guestrin, and M.~Zaharia.
\newblock Semantic operators: A declarative model for rich, ai-based data processing, 2025.
\newblock URL \url{https://arxiv.org/abs/2407.11418}.

\bibitem[Petroni et~al.(2019)Petroni, Rockt{\"{a}}schel, Miller, Lewis, Bakhtin, Wu, and Riedel]{petroni2019language}
F.~Petroni, T.~Rockt{\"{a}}schel, A.~H. Miller, P.~Lewis, A.~Bakhtin, Y.~Wu, and S.~Riedel.
\newblock Language models as knowledge bases?
\newblock In \emph{Proceedings of the 2019 Conference on Empirical Methods in Natural Language Processing (EMNLP)}, 2019.

\bibitem[Petroni et~al.(2020)Petroni, Lewis, Piktus, Rockt{\"a}schel, Wu, Miller, and Riedel]{petroni2020how}
F.~Petroni, P.~Lewis, A.~Piktus, T.~Rockt{\"a}schel, Y.~Wu, A.~H. Miller, and S.~Riedel.
\newblock How context affects language models' factual predictions.
\newblock In \emph{Automated Knowledge Base Construction}, 2020.
\newblock URL \url{https://openreview.net/forum?id=025X0zPfn}.

\bibitem[{Qwen Team}(2025)]{qiao2024qwen2}
{Qwen Team}.
\newblock Qwen2.5 technical report, 2025.
\newblock URL \url{https://arxiv.org/abs/2412.15115}.

\bibitem[Ristad and Yianilos(1998)]{ristad1998learning}
E.~S. Ristad and P.~N. Yianilos.
\newblock Learning string-edit distance.
\newblock \emph{IEEE Transactions on Pattern Analysis and Machine Intelligence}, 20\penalty0 (5):\penalty0 522--532, 1998.

\bibitem[Saeed et~al.(2024)Saeed, Cao, and Papotti]{galois1}
M.~Saeed, N.~D. Cao, and P.~Papotti.
\newblock Querying large language models with {SQL}.
\newblock In \emph{Proceedings 27th International Conference on Extending Database Technology, {EDBT} 2024, Paestum, Italy, March 25 - March 28}, pages 365--372. OpenProceedings.org, 2024.
\newblock \doi{10.48786/EDBT.2024.32}.
\newblock URL \url{https://doi.org/10.48786/edbt.2024.32}.

\bibitem[Saparina and Lapata(2024)]{saparina2024ambrosia}
I.~Saparina and M.~Lapata.
\newblock Ambrosia: A benchmark for parsing ambiguous questions into database queries.
\newblock \emph{Advances in Neural Information Processing Systems}, 37:\penalty0 90600--90628, 2024.

\bibitem[Shankar et~al.(2025)Shankar, Chambers, Shah, Parameswaran, and Wu]{docetl}
S.~Shankar, T.~Chambers, T.~Shah, A.~G. Parameswaran, and E.~Wu.
\newblock Docetl: Agentic query rewriting and evaluation for complex document processing, 2025.
\newblock URL \url{https://arxiv.org/abs/2410.12189}.

\bibitem[Singhal et~al.(2023)Singhal, Azizi, Tu, Mahdavi, Wei, Chung, Scales, Tanwani, Cole-Lewis, Pfohl, Payne, Seneviratne, Gamble, Kelly, Schärli, Chowdhery, Mansfield, y~Arcas, Webster, Corrado, Matias, Chou, Gottweis, Tomasev, Liu, Rajkomar, Barral, Semturs, Karthikesalingam, and Natarajan]{Singhal2023LargeLanguageModelsMedicine}
K.~Singhal, S.~Azizi, T.~Tu, S.~S. Mahdavi, J.~Wei, H.~W. Chung, N.~Scales, A.~Tanwani, H.~Cole-Lewis, S.~Pfohl, P.~Payne, M.~Seneviratne, P.~Gamble, C.~Kelly, N.~Schärli, A.~Chowdhery, P.~Mansfield, B.~A. y~Arcas, D.~Webster, G.~S. Corrado, Y.~Matias, K.~Chou, J.~Gottweis, N.~Tomasev, Y.~Liu, A.~Rajkomar, J.~Barral, C.~Semturs, A.~Karthikesalingam, and V.~Natarajan.
\newblock Large language models encode clinical knowledge.
\newblock \emph{Nature}, 620\penalty0 (7972):\penalty0 172--180, 2023.
\newblock This paper (often associated with Med-PaLM 2) evaluates LLMs on medical benchmarks, highlighting potential but also the need for accuracy and safety.

\bibitem[Stuhler et~al.(0)Stuhler, Ton, and Ollion]{00491241251336794}
O.~Stuhler, C.~D. Ton, and E.~Ollion.
\newblock From codebooks to promptbooks: Extracting information from text with generative large language models.
\newblock \emph{Sociological Methods \& Research}, 0\penalty0 (0):\penalty0 00491241251336794, 0.
\newblock \doi{10.1177/00491241251336794}.
\newblock URL \url{https://journals.sagepub.com/doi/abs/10.1177/00491241251336794}.

\bibitem[Suchanek et~al.(2024)Suchanek, Alam, Bonald, Chen, Paris, and Soria]{suchanek2024yago45largeclean}
F.~Suchanek, M.~Alam, T.~Bonald, L.~Chen, P.-H. Paris, and J.~Soria.
\newblock Yago 4.5: A large and clean knowledge base with a rich taxonomy, 2024.
\newblock URL \url{https://arxiv.org/abs/2308.11884}.

\bibitem[Tang et~al.(2024)Tang, Wang, Qu, Yan, Wu, Zhuang, Kai, Hou, Guo, Zhao, Zhao, and Ma]{tang-etal-2024-itinera}
Y.~Tang, Z.~Wang, A.~Qu, Y.~Yan, Z.~Wu, D.~Zhuang, J.~Kai, K.~Hou, X.~Guo, J.~Zhao, Z.~Zhao, and W.~Ma.
\newblock {I}ti{N}era: Integrating spatial optimization with large language models for open-domain urban itinerary planning.
\newblock In F.~Dernoncourt, D.~Preo{\c{t}}iuc-Pietro, and A.~Shimorina, editors, \emph{Proceedings of the 2024 Conference on Empirical Methods in Natural Language Processing: Industry Track}, pages 1413--1432, Miami, Florida, US, Nov. 2024. Association for Computational Linguistics.
\newblock \doi{10.18653/v1/2024.emnlp-industry.104}.
\newblock URL \url{https://aclanthology.org/2024.emnlp-industry.104/}.

\bibitem[Truhn et~al.(2023)Truhn, Reis-Filho, and Kather]{truhn2023large}
D.~Truhn, J.~S. Reis-Filho, and J.~N. Kather.
\newblock Large language models should be used as scientific reasoning engines, not knowledge databases.
\newblock \emph{Nature medicine}, 29\penalty0 (12):\penalty0 2983--2984, 2023.

\bibitem[Wei et~al.(2024)Wei, Karina, Chung, Jiao, Papay, Glaese, Schulman, and Fedus]{simpleQAOpenAI}
J.~Wei, N.~Karina, H.~W. Chung, Y.~J. Jiao, S.~Papay, A.~Glaese, J.~Schulman, and W.~Fedus.
\newblock Measuring short-form factuality in large language models.
\newblock \emph{CoRR}, abs/2411.04368, 2024.
\newblock \doi{10.48550/ARXIV.2411.04368}.
\newblock URL \url{https://doi.org/10.48550/arXiv.2411.04368}.

\bibitem[Wu et~al.(2025{\natexlab{a}})Wu, Yang, Li, Ji, Okumura, and Zhang]{wu2025mmqa}
J.~Wu, L.~Yang, D.~Li, Y.~Ji, M.~Okumura, and Y.~Zhang.
\newblock {MMQA}: Evaluating {LLM}s with multi-table multi-hop complex questions.
\newblock In \emph{The Thirteenth International Conference on Learning Representations}, 2025{\natexlab{a}}.
\newblock URL \url{https://openreview.net/forum?id=GGlpykXDCa}.

\bibitem[Wu et~al.(2025{\natexlab{b}})Wu, Yang, Chai, Zhang, Liu, Du, Liang, Shu, Cheng, Sun, Li, Li, and Niu]{tableBench2025}
X.~Wu, J.~Yang, L.~Chai, G.~Zhang, J.~Liu, X.~Du, D.~Liang, D.~Shu, X.~Cheng, T.~Sun, T.~Li, Z.~Li, and G.~Niu.
\newblock Tablebench: {A} comprehensive and complex benchmark for table question answering.
\newblock In T.~Walsh, J.~Shah, and Z.~Kolter, editors, \emph{AAAI-25, Sponsored by the Association for the Advancement of Artificial Intelligence, February 25 - March 4, 2025, Philadelphia, PA, {USA}}, pages 25497--25506. {AAAI} Press, 2025{\natexlab{b}}.
\newblock \doi{10.1609/AAAI.V39I24.34739}.
\newblock URL \url{https://doi.org/10.1609/aaai.v39i24.34739}.

\bibitem[Yu et~al.(2018)Yu, Zhang, Yang, Yasunaga, Wang, Li, Ma, Li, Yao, Roman, et~al.]{yu2018spider}
T.~Yu, R.~Zhang, K.~Yang, M.~Yasunaga, D.~Wang, Z.~Li, J.~Ma, I.~Li, Q.~Yao, S.~Roman, et~al.
\newblock Spider: A large-scale human-labeled dataset for complex and cross-domain semantic parsing and text-to-sql task.
\newblock In \emph{2018 Conference on Empirical Methods in Natural Language Processing, EMNLP 2018}, pages 3911--3921. Association for Computational Linguistics, 2018.

\bibitem[Zhang et~al.(2025)Zhang, Luo, Zhang, Ma, Zhang, Li, Li, Yao, Xu, Zhou, Zhang-Li, Yu, Zhao, Li, and Tang]{zhang2025tablellmenablingtabulardata}
X.~Zhang, S.~Luo, B.~Zhang, Z.~Ma, J.~Zhang, Y.~Li, G.~Li, Z.~Yao, K.~Xu, J.~Zhou, D.~Zhang-Li, J.~Yu, S.~Zhao, J.~Li, and J.~Tang.
\newblock Tablellm: Enabling tabular data manipulation by llms in real office usage scenarios, 2025.
\newblock URL \url{https://arxiv.org/abs/2403.19318}.

\bibitem[Zhong et~al.(2017)Zhong, Xiong, and Socher]{Zhong2017Seq2SQL}
V.~Zhong, C.~Xiong, and R.~Socher.
\newblock Seq2sql: Generating structured queries from natural language using reinforcement learning.
\newblock \emph{arXiv preprint arXiv:1709.00103}, 2017.

\bibitem[Zhu et~al.(2021)Zhu, Lei, Huang, Wang, Zhang, Lv, Feng, and Chua]{zhu2021tat}
F.~Zhu, W.~Lei, Y.~Huang, C.~Wang, S.~Zhang, J.~Lv, F.~Feng, and T.-S. Chua.
\newblock {TAT}-{QA}: A question answering benchmark on a hybrid of tabular and textual content in finance.
\newblock In \emph{Proceedings of the 59th Annual Meeting of the Association for Computational Linguistics and the 11th International Joint Conference on Natural Language Processing (Volume 1: Long Papers)}, pages 3277--3287, Online, Aug. 2021. Association for Computational Linguistics.
\newblock \doi{10.18653/v1/2021.acl-long.254}.
\newblock URL \url{https://aclanthology.org/2021.acl-long.254}.

\end{thebibliography}

\newpage
\appendix

\section{Motivation Example}

As discussed in the introduction, extracting structured information from an LLM's internal knowledge in a tabular format poses unique and challenging problems, distinct from conventional, single-point natural language queries. To demonstrate and isolate these issues empirically, we designed a series of controlled experiments.

Specifically, we manually curated a compact yet informative dataset that includes detailed statistics for the 23 players who represented the Italian national football team at UEFA Euro 2016. For each player, the dataset contains their surname, date of birth, and jersey number used in the tournament. Additionally, it includes six per-season attributes (club, appearances, goals, assists, yellow cards, red cards) across nine different seasons. The final dataset thus comprises 23 rows and 57 attributes. All this information is publicly available (e.g., on Wikipedia) and is assumed to be part of the internal knowledge of the tested LLMs.

Unlike the broader evaluation presented in Section~\ref{sec:exp}, the aim of these specific experiments is not to assess overall extraction accuracy, but rather to evaluate how performance degrades due to conditions inherent to tabular queries.

\paragraph{Experiment 1: Incremental Attribute Request.}
In the first experiment, we progressively increased the number of requested attributes, from 1 to 57. In the first iteration, the model was asked to return only the surname of the 23 Italian Euro 2016 players. In the second, both surname and date of birth were requested, and so on, up to all 57 attributes. Crucially, after each query, the F1 score was computed solely on the \textit{surname} column. The goal was to assess whether the accuracy of a fixed attribute degrades as the number of requested attributes increases. Ideally, the quality on \textit{surname} should remain constant regardless of how many other attributes are queried.

However, as shown in Figure~\ref{fig:tabular}.(a), performance on the \textit{surname} column degrades substantially. For instance, using GPT-4.1, quality drops from 1.0 when a single attribute is requested to 0.516 when 53 attributes are included. This experiment was conducted using the \textsc{CoT} prompting strategy, which yielded the best results in our main benchmark. Similar trends were observed across other LLMs and prompting strategies. These findings support the hypothesis that tabular queries induce specific degradation patterns that are not typically observed in more natural, conversational settings.

\begin{figure}
    \begin{minipage}[t]{0.5\columnwidth}
        \centering
        \resizebox{1.0\columnwidth}{!}{%
            \begin{tikzpicture}
              \begin{axis}[
                xlabel={\# Number of requested attributes},
                ylabel={F1 Score (surname)},
                grid=major,
                ymin=0.0, ymax=1,
                xmin=1.0, xmax=57.0,
                width=11.1cm,
                height=8cm,
                enlargelimits=false,
                legend style={
                  at={(0.5,1)}, 
                  anchor=south,    
                  legend columns=4,
                  fill=white,
                  draw=none,
                  align=left
                },
                cycle multiindex* list={
                  color list
                },
                clip=false 
              ]
          
\addplot+[line width=1.64pt] coordinates { (1, 0.939) (2, 0.936) (3, 0.913) (4, 0.957) (5, 0.894) (6, 0.837) (7, 0.87) (8, 0.857) (9, 0.889) (10, 0.875) (11, 0.913) (12, 0.829) (13, 0.905) (14, 0.909) (15, 0.889) (16, 0.629) (17, 0.889) (18, 0.78) (19, 0.649) (20, 0.737) (21, 0.844) (22, 0.8) (23, 0.844) (24, 0.737) (25, 0.629) (26, 0.857) (27, 0.667) (28, 0.703) (29, 0.889) (30, 0.7) (31, 0.78) (32, 0.75) (33, 0.75) (34, 0.791) (35, 0.588) (36, 0.595) (37, 0.75) (38, 0.78) (39, 0.718) (40, 0.81) (41, 0.571) (42, 0.667) (43, 0.611) (44, 0.737) (45, 0.78) (46, 0.588) (47, 0.588) (48, 0.588) (49, 0.588) (50, 0.588) (51, 0.667) (52, 0.562) (53, 0.588) (54, 0.686) (55, 0.588) (56, 0.516) (57, 0.516) };
\addlegendentry{LLama3.3-70B}  
\addplot+[line width=1.64pt] coordinates {(1, 0.476) (2, 0.702) (3, 0.711) (4, 0.679) (5, 0.682) (6, 0.717) (7, 0.691) (8, 0.711) (9, 0.711) (10, 0.565) (11, 0.727) (12, 0.667) (13, 0.667) (14, 0.714) (15, 0.683) (16, 0.682) (17, 0.718) (18, 0.681) (19, 0.667) (20, 0.634) (21, 0.615) (22, 0.703) (23, 0.722) (24, 0.703) (25, 0.7) (26, 0.684) (27, 0.5) (28, 0.667) (29, 0.516) (30, 0.649) (31, 0.686) (32, 0.737) (33, 0.467) (34, 0.467) (35, 0.629) (36, 0.686) (37, 0.629) (38, 0.629) (39, 0.414) (40, 0.571) (41, 0.571) (42, 0.485) (43, 0.545) (44, 0.611) (45, 0.545) (46, 0.545) (47, 0.562) (48, 0.296) (49, 0.414) (50, 0.296) (51, 0.296) (52, 0.296) (53, 0.296) (54, 0.562) (55, 0.296) (56, 0.452) (57, 0.452)  };
\addlegendentry{LLama3.1-8B}
\addplot+[line width=1.64pt] coordinates {(1, 0.979) (2, 1.000) (3, 1.000) (4, 0.955) (5, 0.978) (6, 0.957) (7, 0.955) (8, 0.93) (9, 0.909) (10, 1) (11, 0.978) (12, 0.93) (13, 0.978) (14, 0.955) (15, 0.606) (16, 0.606) (17, 0.606) (18, 0.606) (19, 0.606) (20, 0.606) (21, 0.757) (22, 0.606) (23, 0.606) (24, 0.606) (25, 0.606) (26, 0.647) (27, 0.647) (28, 0.606) (29, 0.85) (30, 0.606) (31, 0.878) (32, 0.516) (33, 0.722) (34, 0.588) (35, 0.516) (36, 0.647) (37, 0.789) (38, 0.647) (39, 0.686) (40, 0.357) (41, 0.647) (42, 0.414) (43, 0.516) (44, 0.562) (45, 0.562) (46, 0.606) (47, 0.467) (48, 0.562) (49, 0.905) (50, 0.686) (51, 0.606) (52, 0.562) (53, 0.562) (54, 0.606) (55, 0.516) (56, 0.606) (57, 0.647)};
\addlegendentry{GPT-4.1}
\addplot+[line width=1.64pt] coordinates {(1, 0.723) (2, 0.727) (3, 0.682) (4, 0.762) (5, 0.714) (6, 0.829) (7, 0.545) (8, 0.485) (9, 0.588) (10, 0.5) (11, 0.485) (12, 0.562) (13, 0.516) (14, 0.545) (15, 0.545) (16, 0.467) (17, 0.357) (18, 0.357) (19, 0.231) (20, 0.357) (21, 0.357) (22, 0.357) (23, 0.357) (24, 0.357) (25, 0.357) (26, 0.357) (27, 0.357) (28, 0.357) (29, 0.296) (30, 0.296) (31, 0.231) (32, 0.231) (33, 0.286) (34, 0.231) (35, 0.357) (36, 0.357) (37, 0.231) (38, 0.357) (39, 0.231) (40, 0.231) (41, 0.231) (42, 0.231) (43, 0.4) (44, 0.231) (45, 0.231) (46, 0.231) (47, 0.231) (48, 0.296) (49, 0.286) (50, 0.231) (51, 0.231) (52, 0.16) (53, 0.231) (54, 0.231) (55, 0.231) (56, 0.231) (57, 0.16) };
\addlegendentry{GPT-4.1-mini}
            
              \end{axis}
            \end{tikzpicture}
        }
        \caption*{(a) Experiment 1: Incremental Attribute Request.}
        \label{fig:tabular-a}
    \end{minipage}
    \begin{minipage}[t]{0.5\columnwidth}
        \centering
        \resizebox{1.0\columnwidth}{!}{%
            \begin{tikzpicture}
              \begin{axis}[
    xlabel={Position of the 'surname' attribute},
    ylabel={F1 Score (surname)},
    grid=major,
    ymin=0.0, ymax=1,
    xmin=1.0, xmax=30.0,
    width=11.1cm,
    height=8cm,
    enlargelimits=false,
    legend style={
      at={(0.5,1)}, 
      anchor=south,    
      legend columns=4,
      fill=white,
      draw=none,
      align=left
    },
    cycle multiindex* list={
      color list
    },
    clip=false 
  ]
     
\addplot+[line width=1.64pt] coordinates {(1, 0.7) (2, 0.913) (3, 0.913) (4, 0.615) (5, 0.343) (6, 0.696) (7, 0.615) (8, 0.579) (9, 0.711) (10, 0.533) (11, 0.432) (12, 0.564) (13, 0.541) (14, 0.457) (15, 0.286) (16, 0.489) (17, 0.439) (18, 0.522) (19, 0.533) (20, 0.359) (21, 0.368) (22, 0.478) (23, 0.304) (24, 0.435) (25, 0.391) (26, 0.486) (27, 0.457) (28, 0.462) (29, 0.435) (30, 0.522)  };
\addlegendentry{LLama3.3-70B}       
\addplot+[line width=1.64pt] coordinates {(1, 0.649) (2, 0.615) (3, 0.462) (4, 0.462) (5, 0.359) (6, 0.256) (7, 0.462) (8, 0.41) (9, 0.308) (10, 0.308) (11, 0.171) (12, 0.462) (13, 0.205) (14, 0.308) (15, 0.308) (16, 0.205) (17, 0.368) (18, 0.263) (19, 0.564) (20, 0.359) (21, 0.171) (22, 0.316) (23, 0.263) (24, 0.368) (25, 0.308) (26, 0.564) (27, 0.316) (28, 0.632) (29, 0.359) (30, 0.564)  };
\addlegendentry{LLama3.1-8B}
\addplot+[line width=1.64pt] coordinates {(1, 0.606) (2, 0.737) (3, 0.647) (4, 0.545) (5, 0.485) (6, 0.345) (7, 0.595) (8, 0.345) (9, 0.545) (10, 0.387) (11, 0.424) (12, 0.286) (13, 0.588) (14, 0.5) (15, 0.5) (16, 0.5) (17, 0.485) (18, 0.438) (19, 0.424) (20, 0.323) (21, 0.4) (22, 0.286) (23, 0.364) (24, 0.387) (25, 0.424) (26, 0.424) (27, 0.545) (28, 0.424) (29, 0.387) (30, 0.5) };
\addlegendentry{GPT-4.1}
\addplot+[line width=1.64pt] coordinates {(1, 0.296) (2, 0.231) (3, 0.286) (4, 0.077) (5, 0.231) (6, 0.154) (7, 0.222) (8, 0.077) (9, 0.077) (10, 0.077) (11, 0.077) (12, 0.077) (13, 0.154) (14, 0) (15, 0.077) (16, 0.077) (17, 0.077) (18, 0) (19, 0.077) (20, 0) (21, 0.077) (22, 0.077) (23, 0.08) (24, 0.074) (25, 0.077) (26, 0.077) (27, 0.154) (28, 0.154) (29, 0.077) (30, 0) };
\addlegendentry{GPT-4.1-mini}

              \end{axis}
            \end{tikzpicture}
        }
        \caption*{(b) Experiment 2: Attribute Position Sensitivity.}
        \label{fig:tabular-b}
    \end{minipage}
\caption{F1 Quality of attribute \textit{surname} in different queries}
\label{fig:tabular}
\end{figure}

\paragraph{Experiment 2: Attribute Position Sensitivity.}
The second experiment evaluated whether the \textit{position} of an attribute in the output affects its quality. A fixed query requesting 30 attributes was used as a base. We then generated 30 query variants, each placing the \textit{surname} attribute at a different position (from 1st to 30th). Again, F1 scores were computed only on the \textit{surname} column.

As illustrated in Figure~\ref{fig:tabular}.(b), the position of the attribute has a significant impact. Unlike the previous experiment, the size of the output remains constant across all variants; the only variable is the position of the \textit{surname} attribute. Maximum accuracy is achieved when the attribute appears first, with performance declining as the attribute shifts toward the middle positions. 

We repeated the same experiments for the running example introduced in Section 1, the US Counties dataset, which has a small number of attributes. We first measured the F1 score using only the $<$County, State$>$ attribute pair. However, when we added the Area attribute to the returned attributes in the query, we observed a significant drop in F1. Most of the previously correct $<$County, State$>$ pairs were no longer returned. For example, only 51 out of 3246 expected tuples were retrieved, and 3 of them included hallucinated $<$County, State$>$ combinations. Further inspection revealed that 40 of the 51 returned tuples also contained incorrect values for the Area attribute.

These preliminary experiments highlight structural phenomena that are intrinsic to tabular-style querying. They underline how even high-performing LLMs can suffer from specific degradation modes when asked to return structured multi-attribute outputs, making this task significantly more challenging than conventional QA scenarios.

\section{More Details on RelationalFactQA Dataset}

\stitle{RFQA Detailed Statistics.}
Table~\ref{tab:data_summary} presents detailed statistics of the RFQA dataset. It reports the minimum, maximum, first quartile (Q1), third quartile (Q3), and average values for the expected number of output tuples, attributes, and cells. Additionally, the attribute dimension is further analyzed by type, providing the same set of statistics separately for numerical attributes, categorical attributes, and mixed attributes (containing both numerical and categorical values).

The training set statistics of the RFQA dataset reveal a highly skewed and diverse structure. The average number of output tuples per instance is about 27, but the maximum value reaches 904, indicating that while most cases are relatively small, there are some significantly larger ones, pointing to a long-tailed distribution. Similarly, the number of output cells ranges from 1 to 4500, with an average of 135.5, suggesting considerable variation in instance complexity. On the attribute side, most examples contain a mix of categorical (avg. 3.16) and mixed attributes (avg. 4.26), with relatively few numerical attributes (avg. 1.06). This implies that RFQA poses both relational and interpretative challenges, as models must handle heterogeneous data types and cope with a wide range of input sizes.

\begin{table}[htbp] 
  \centering
  \caption{Statistics of \textbf{RFQA}.}
  \label{tab:data_summary}
  \begin{tabular}{l|lll|lll} 
    {} & \textbf{Output} & \textbf{Output} & \textbf{Output} & \textbf{Attribute} & \textbf{Attribute} & \textbf{Attribute} \\
    \textbf{Dimension} & \textbf{Tuple} & \textbf{Attributes} & \textbf{Cells} & \textbf{Numerical} & \textbf{Categorical} & \textbf{Mixed}\\
    \hline
    MIN & 1 & 1 & 1 & 1 & 1 & 2 \\
    Q1 & 1 & 2 & 3 & 1 & 2 & 5 \\
    AVG & 26.94 & 5.32 & 135.50 & 1.06 & 3.16 & 4.26 \\
    Q3 & 8 & 9 & 45 & 3 & 6 & 9 \\
    MAX & 904 & 9 & 4500 & 4 & 6 & 9 \\
    \hline
  \end{tabular}
\end{table}

Figure~\ref{fig:topicDistribution} shows the distribution of topics covered by the queries in the dataset, spanning 27 distinct categories. The most frequent topics are related to automatically generated queries, including Nobel Prize winners, chemical elements from the periodic table, popular web search engines, video game publishers, and global airports. Additional topics are sourced from the Spider and Bird corpora.

\begin{figure}[th]
    \centering
\includegraphics[width=0.6\columnwidth]{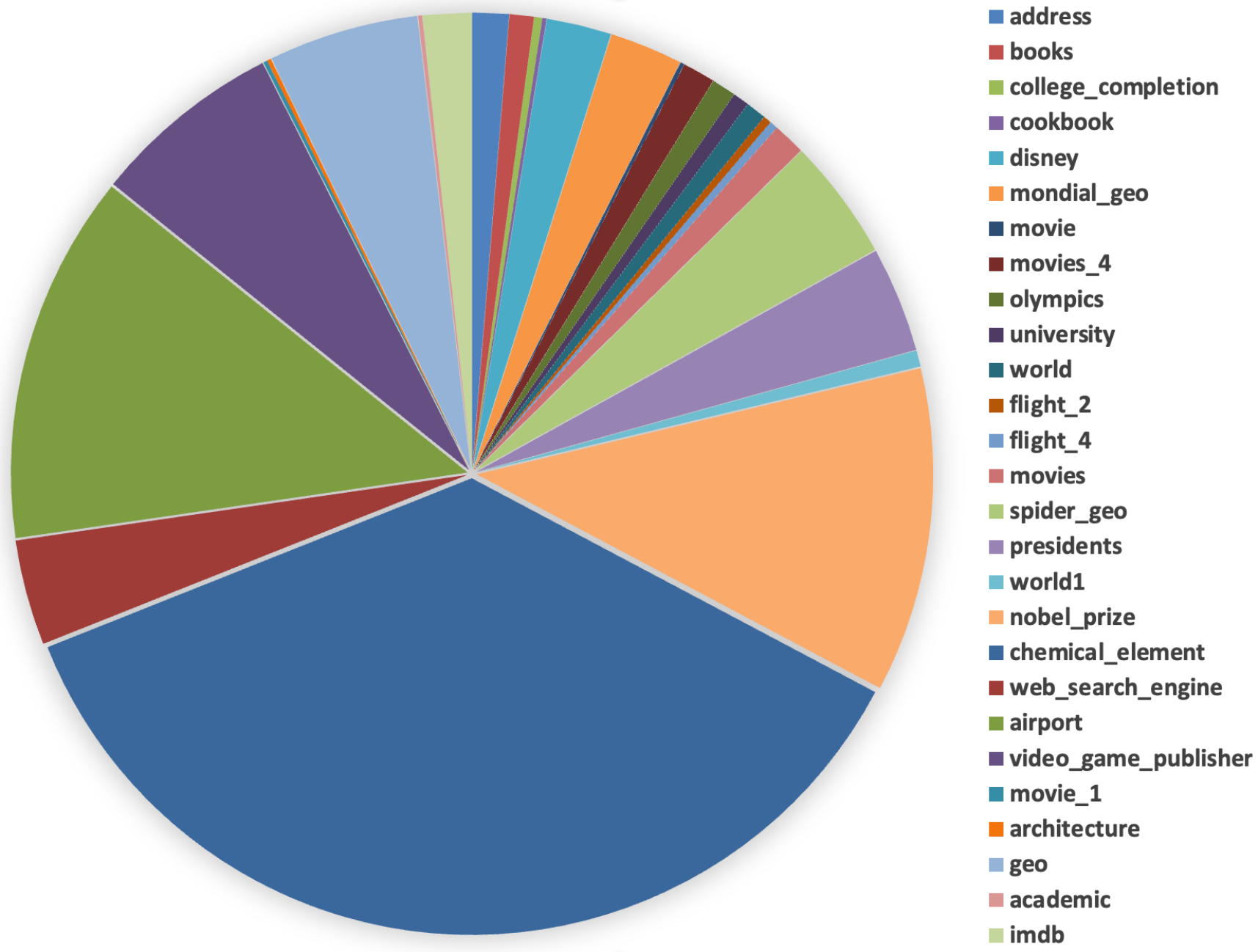}
    \caption{Topic distribution}
    \label{fig:topicDistribution}
\end{figure}

\stitle{Metadata.}
RFQA also includes rich metadata that can be used to analyze queries in terms of their expected output size (rows, attributes, cells), the nature of the selected attributes (whether numerical, categorical, or mixed), and their complexity (e.g., presence of \textsc{WHERE} conditions involving numerical or categorical filters). Below, we describe the fields available in the RFQA dataset:
\begin{itemize}[leftmargin=*]
\item \textsc{q\_id}: Unique identifier for each query instance.
\item \textsc{dataset}: Name of the source corpus. Possible values are: ``bird'', ``galois'', ``spider1'', and ``qatch''. The first three represent queries from existing corpora, while ``qatch'' refers to automatically generated queries.
\item \textsc{db\_id}: Identifier of the associated database (i.e., the topic) for each query.
\item \textsc{sql}: SQL query in PostgreSQL syntax.
\item \textsc{question}: Natural language (NL) version of the query.
\item \textsc{tuples}: Expected number of output tuples (rows).
\item \textsc{attrs}: Expected number of output attributes (columns).
\item \textsc{attr numerical}: Number of expected numerical attributes.
\item \textsc{attr categorical}: Number of expected categorical attributes.
\item \textsc{tables}: Number of tables referenced in the \textsc{FROM} clause.
\item \textsc{numerical conditions}: Number of numerical conditions in the \textsc{WHERE} clause.
\item \textsc{categorical conditions}: Number of categorical conditions in the \textsc{WHERE} clause.
\item \textsc{aggr}: Set to 1 if the query includes an aggregation function, 0 otherwise.
\item \textsc{join}: Number of joins in the query.
\item \textsc{distinct}: Set to 1 if the query includes the \textsc{DISTINCT} operator, 0 otherwise.
\item \textsc{group by}: Set to 1 if the query includes the \textsc{GROUP BY} operator, 0 otherwise.
\item \textsc{limit}: Set to 1 if the query includes the \textsc{LIMIT} operator, 0 otherwise.
\item \textsc{order by}: Set to 1 if the query includes the \textsc{ORDER BY} operator, 0 otherwise.
\item \textsc{attr num \& cat}: Sum of numerical and categorical attributes in the \textsc{SELECT} clause. Present only if both types are used.
\item \textsc{con num \& cat}: Sum of numerical and categorical conditions in the \textsc{WHERE} clause. Present only if both types are used.
\item \textsc{cells}: Expected number of cells (i.e., total elements in the output table).
\item \textsc{backlinks}: Number of backlinks retrieved using the Wikipedia API.
\end{itemize}

The actual data for each NL-SQL query pair is stored in the data folder, following the structure \texttt{<dataset>/<db\_id>}. To retrieve the expected results for a given query $q$, one can execute the corresponding SQL query on its associated database that can be loaded with the associated data. In our experiment all the data are imported into PostgreSQL database.

The backlinks for each query are calculated as follows: for each row of a table $t$ result of a query $q$, the key value of the table is used to search for a relative Wikipedia page referencing the entity described by the current row. If the search is successful, the backlinks are directly extracted from the page using the Wikipedia API. The total number of backlinks for a given query is then computed as the average number of backlinks of the single query results.

\section{Prompts and LLM Response Processing}
The prompting strategy used is iterative and consists of two main steps: a \textit{start prompt}, which instructs the LLM on the type of data to extract, and an \textit{iterative prompt}, which guides the model to retrieve additional data if more is available. Figure~\ref{fig:nlPrompt} shows the prompts used for the natural language (NL) strategy; Figure~\ref{fig:sqlPrompt} presents the prompt for the SQL-based strategy; and Figure~\ref{fig:cotPrompt} illustrates the prompt used for the Chain-of-Thought (CoT) strategy.

\begin{figure}[h!]
    \centering
    \resizebox{1.0\textwidth}{!}{%
    \noindent\fbox{%
    \parbox{0.98\columnwidth}{%
        \textbf{Start Prompt:} \textit{NL Question}. Respond with JSON only. Don't add any comments. Use the following JSON schema: \textit{jsonSchema}.\\
        \\
        \textit{NL Question:} is the query in natural language\\
        \textit{jsonSchema:} is the schema of the tabular response translated in JSON schema\\
        \\
        \textbf{Iterative Prompt:} List more values if there are more, otherwise return an empty JSON. Respond with JSON only.       
    }%
}
}
\caption{NL Prompt Syntax. Text in \textit{italic} is injected from the given NL query and the expected JSON schema of the response.}
\label{fig:nlPrompt}
\end{figure}

\begin{figure}[h!]
    \centering
    \resizebox{1.0\textwidth}{!}{%
    \noindent\fbox{%
    \parbox{0.98\columnwidth}{%
        \textbf{Start Prompt:} List the results of the SQL query: \textit{SQL}. Respond with JSON only. Don't add any comments. Use the following JSON schema: \textit{jsonSchema}.\\
        \\
        \textit{SQL:} is the query in SQL syntax\\
        \textit{jsonSchema:} is the schema of the tabular response translated in JSON schema\\
        \\
        \textbf{Iterative Prompt:} List more values if there are more, otherwise return an empty JSON. Respond with JSON only.       
    }%
}
}
\caption{SQL Prompt Syntax. Text in \textit{italic} is injected from the given SQL query and the expected JSON schema of the response.}
\label{fig:sqlPrompt}
\end{figure}

\begin{figure}[h!]
    \centering
    \resizebox{1.0\textwidth}{!}{%
    \noindent\fbox{%
    \parbox{0.98\columnwidth}{%
        \textbf{First Prompt:} Given the following query, populate the table with actual values.
query: select \textit{attributes} from \textit{table} (where \textit{conditions}). Respond with JSON only. Don't add any comments. Use the following JSON schema: \textit{jsonSchema}.\\
        \\
        \textit{attributes:} is the set of attribute names of the table \textit{table}\\
        \textit{table:} is the table name\\
        \textit{conditions:} the condition(s) if passed\\
        \textit{jsonSchema:} is the schema of the table translated in JSON schema\\
        \\
        \textbf{Iterative Prompt:} List more values if there are more, otherwise return an empty JSON. Respond with JSON only.       
    }%
}
}
\caption{CoT Prompt Syntax. Text in \textit{italic} is injected from the given SQL query. Values between parenthesis are populated only if the condition(s) is given.}
\label{fig:cotPrompt}
\end{figure}

\textbf{Handling Output JSON Errors}. All the strategies ask the LLM to return the data in a structured form respecting a JSON format prompted. We parse the response according to the required JSON. The most common issues in the JSON parsing and our corresponding handling methods are the following:
\begin{itemize}[leftmargin=*]
    \item \textit{Malformed JSON syntax}: This includes missing quotation marks or improperly formatted numbers. In such cases, we re-prompt the LLM, explicitly asking it to return the answer in valid JSON format.
    \item \textit{Truncated or broken JSON}: Often caused by the model hitting its maximum token limit. When this happens, we identify the last unmatched opening brace and extract the content up to that point. We then attempt to complete the JSON structure by adding the necessary closing braces to recover a valid object.
\end{itemize}
If none of the recovery strategies succeed, we terminate the iteration and treat the response as invalid.

\section{Details on The Experiments}
\subsection{Used Models}
Table~\ref{tab:models} lists the models used in our evaluation. For each model, we provide its full name, along with the version or release date when available. All models are accessed via their respective APIs.

\begin{table}[h!]
\centering
\begin{tabular}{@{}lll@{}}
\toprule
\textbf{Model Name} & \textbf{Model Full Name} & \textbf{Platform} \\ \midrule
GPT 4.1 & gpt-4.1-2025-04-14 & OpenAI \\
GPT 4.1 mini & gpt-4.1-mini-2025-04-14 & OpenAI \\
Mistral 7B & Mistral (7B) Instruct v0.3 Released May 22, 2024 & Together AI \\
QWEN 2.5-7B & Qwen2.5 7B Instruct Turbo * & Together AI \\
LLama 3.1-8B & Meta Llama 3.1 8B Instruct Turbo * & Together AI \\
LLama 3.3 70B & Meta Llama 3.3 70B Instruct Turbo * & Together AI \\
Gemma 2-9B & Gemma-2 Instruct (9B) & Together AI \\
DeepSeek 70B & DeepSeek R1 Distill Llama 70B Released Jan 20, 2025 & Together AI \\
QWEN 3-235B & Qwen3 235B A22B FP8 Throughput Released Apr 27, 2025 & Together AI \\ \bottomrule
\end{tabular}
\caption{Overview of used Models and the respective Platforms. * indicates that there is no release date}
\label{tab:models}
\end{table}

\subsection{Additional Results}
Table~\ref{tab:benchmarkResultsDetailes} reports the detailed results for Exp-1 (Overall Performance). It extends results reported in Table~\ref{tab:benchmarkResults} by also adding the Precision and Recall computed at the cell level. In the following we concentrate on the Precision and Recall to also motivate the F1 computed in Exp-1 (Overall Performance). 

Table~\ref{tab:benchmarkResultsDetailes} reveal notable trends, particularly in Precision (P) and Recall (R) across different prompting strategies (NL, SQL, CoT) and LLMs of varying sizes. Overall, the Chain-of-Thought (CoT) strategy consistently outperforms the NL and SQL settings in both precision and recall, especially for larger models like GPT 4.1, QWEN 3-235B. These models demonstrate the strongest balance between precision and recall, with GPT 4.1 achieving the highest CoT precision (0.720) and a strong recall (0.691), reflecting its ability to generate accurate and comprehensive responses. Smaller models, such as Mistral 7B and LLama 3.1-8B, show more variability, with generally lower precision in SQL and NL strategies and modest gains under CoT prompting. Interestingly, certain mid-sized models like Gemma 2-9B, LLama 3.3-70B and DeepSeek 70B achieve competitive performance, indicating that parameter count alone is not the sole determinant of quality. In general, the CoT strategy enhances both recall and precision, suggesting that structured reasoning prompts help LLMs capture more relevant data points while maintaining correctness. 

\setlength{\tabcolsep}{.45em}

\begin{table}[h!] 
\small
  \centering
  \caption{Benchmark Complete Results. Precision (P), Recall (R), F1 and Tuple Similarity (TS) measured for all LLMs in our evaluation. AVG is the average between F1 and TS. LLMs ordered by increasing size in terms of parameters.}
  \label{tab:benchmarkResultsDetailes}
\begin{tabular}{clccccccccc}
\hline
\textbf{} & \textbf{} & 
\shortstack{Mistral \\ 7B} & 
\shortstack{\rule{0pt}{2.2ex} QWEN \\ 2.5-7B} & 
\shortstack{LLama \\ 3.1-8B} & 
\shortstack{GPT \\ 4.1 mini} & 
\shortstack{Gemma \\ 2-9B} & 
\shortstack{LLama \\ 3.3-70B} & 
\shortstack{DeepSeek \\ 70B} & 
\shortstack{GPT \\ 4.1} & 
\shortstack{QWEN \\ 3-235B} \\
\hline

\multirow{5}{*}{\textbf{NL}} 
& P & \colorcell{0.459} & \colorcell{0.546} & \colorcell{0.473} & \colorcell{0.593} & \colorcell{0.596} & \colorcell{0.608} & \colorcell{0.649} & \colorcell{0.682} & \colorcell{0.635} \\
& R & \colorcell{0.477} & \colorcell{0.477} & \colorcell{0.608} & \colorcell{0.525} & \colorcell{0.556} & \colorcell{0.657} & \colorcell{0.615} & \colorcell{0.652} & \colorcell{0.618} \\
& F1 & \colorcell{0.44} & \colorcell{0.487} & \colorcell{0.481} & \colorcell{0.537} & \colorcell{0.557} & \colorcell{0.609} & \colorcell{0.606} & \colorcell{0.654} & \colorcell{0.613} \\
& TS & \colorcell{0.076} & \colorcell{0.085} & \colorcell{0.155} & \colorcell{0.115} & \colorcell{0.107} & \colorcell{0.149} & \colorcell{0.15} & \colorcell{0.247} & \colorcell{0.225} \\
& AVG & \colorcell{0.258} & \colorcell{0.286} & \colorcell{0.318} & \colorcell{0.326} & \colorcell{0.332} & \colorcell{0.379} & \colorcell{0.378} & \colorcell{0.45} & \colorcell{0.419} \\
\hline
\multirow{5}{*}{\textbf{SQL}} 
& P & \colorcell{0.354} & \colorcell{0.537} & \colorcell{0.327} & \colorcell{0.457} & \colorcell{0.62} & \colorcell{0.626} & \colorcell{0.664} & \colorcell{0.401} & \colorcell{0.642} \\
& R & \colorcell{0.423} & \colorcell{0.451} & \colorcell{0.438} & \colorcell{0.409} & \colorcell{0.566} & \colorcell{0.651} & \colorcell{0.586} & \colorcell{0.391} & \colorcell{0.587} \\
& F1 & \colorcell{0.346} & \colorcell{0.459} & \colorcell{0.332} & \colorcell{0.417} & \colorcell{0.571} & \colorcell{0.62} & \colorcell{0.6} & \colorcell{0.388} & \colorcell{0.595} \\
& TS & \colorcell{0.042} & \colorcell{0.079} & \colorcell{0.11} & \colorcell{0.055} & \colorcell{0.123} & \colorcell{0.155} & \colorcell{0.142} & \colorcell{0.096} & \colorcell{0.185} \\
& AVG & \colorcell{0.194} & \colorcell{0.269} & \colorcell{0.221} & \colorcell{0.236} & \colorcell{0.347} & \colorcell{0.387} & \colorcell{0.371} & \colorcell{0.242} & \colorcell{0.39} \\
\hline
\multirow{5}{*}{\textbf{CoT}} 
& P & \colorcell{0.544} & \colorcell{0.6} & \colorcell{0.593} & \colorcell{0.693} & \colorcell{0.645} & \colorcell{0.686} & \colorcell{0.696} & \colorcell{0.72} & \colorcell{0.692} \\
& R & \colorcell{0.469} & \colorcell{0.48} & \colorcell{0.614} & \colorcell{0.626} & \colorcell{0.578} & \colorcell{0.701} & \colorcell{0.634} & \colorcell{0.691} & \colorcell{0.641} \\
& F1 & \colorcell{0.477} & \colorcell{0.503} & \colorcell{0.585} & \colorcell{0.638} & \colorcell{0.594} & \colorcell{0.677} & \colorcell{0.646} & \colorcell{0.693} & \colorcell{0.651} \\
& TS & \colorcell{0.09} & \colorcell{0.091} & \colorcell{0.127} & \colorcell{0.12} & \colorcell{0.106} & \colorcell{0.157} & \colorcell{0.168} & \colorcell{0.174} & \colorcell{0.228} \\
& AVG & \colorcell{0.284} & \colorcell{0.297} & \colorcell{0.356} & \colorcell{0.379} & \colorcell{0.35} & \colorcell{0.417} & \colorcell{0.407} & \colorcell{0.433} & \colorcell{0.439} \\
\hline
\end{tabular}
\end{table}

Figures~\ref{fig:impactAttrsTSFull} and~\ref{fig:impactAttrsF1Full} contain all the tested LLM performances based on the number of the attributes requested in the query respectively on measures against the TS metric (the previous) and the F1 metric (the latter).

The results indicate that Tuple Similarity (TS) generally decreases as the number of attributes increases beyond three, across most models and prompting strategies. Natural Language (NL) shows peak TS performance at low attribute counts (around two to three), with a sharp decline afterward. Models such as QWEN 3 and GPT-4.1 perform well on TS with fewer attributes but degrade quickly with increased complexity. Conversely, F1 scores tend to be more stable or even improve with more attributes, especially for models like Llama 3.3 and Gemma 2, suggesting these models handle complex outputs better in terms of generation quality. SQL prompting exhibits lower TS overall compared to NL, with early peaks and rapid decline; however, models like Llama 3.3 and DeepSeek demonstrate relative robustness at moderate attribute counts. Chain-of-Thought (CoT) prompting shows improved TS retention for GPT-4.1 and Llama 3.3 as attribute numbers rise, highlighting the advantage of step-by-step reasoning for managing complexity. Additionally, CoT achieves the highest F1 scores overall, particularly with reasoning-focused models maintaining strong performance regardless of attribute count. Across all experiments, Llama 3.3 consistently outperforms other models in both TS and F1 metrics, especially as task complexity grows, while Gemma 2 and GPT-4.1 remain competitive in F1 but are more sensitive in TS. QWEN 3 exhibits inconsistent TS results despite some strength in F1. These observations underline the importance of both prompt strategy and model choice in handling increasing task complexity.

\begin{figure}
    \begin{minipage}[h]{0.33\columnwidth}
        \centering
        \resizebox{1.0\columnwidth}{!}{%
            \begin{tikzpicture}
              \begin{axis}[
                    xlabel={\# Attrs.},
                    grid=major,
                    ymin=0.0, ymax=0.75,
                    xmin=1.0, xmax=9.0,
                    legend style = {at={(0.23,1.0)}, legend columns=3, anchor=north west, fill=white,align=left},
                    width=11.1cm,
                    height=8cm,
                    cycle multiindex* list={
                      mark list*\nextlist
                      color list
                    }
                  ]
            
                \addplot+[line width=1.64pt] coordinates {(1, 0.18) (2, 0.204) (3, 0.236) (4, 0.147) (5, 0.004) (6, 0.0) (9, 0.0)}; 
                \addlegendentry{Mistral}
                \addplot+[line width=1.64pt] coordinates {(1, 0.195) (2, 0.221) (3, 0.276) (4, 0.059) (5, 0.002) (6, 0.0) (9, 0.002)}; 
                \addlegendentry{Qwen 2.5}
                \addplot+[line width=1.64pt] coordinates {(1, 0.328) (2, 0.393) (3, 0.37) (4, 0.117) (5, 0.078) (6, 0.005) (9, 0.036)}; 
                \addlegendentry{Llama 3.1}
                \addplot+[line width=1.64pt] coordinates {(1, 0.27) (2, 0.225) (3, 0.135) (4, 0.125) (5, 0.041) (6, 0.054) (9, 0.046)}; 
                \addlegendentry{GPT-4.1 mini}
                \addplot+[line width=1.64pt] coordinates {(1, 0.297) (2, 0.243) (3, 0.238) (4, 0.059) (5, 0.007) (6, 0.004) (9, 0.004)}; 
                \addlegendentry{Gemma 2}
                \addplot+[line width=1.64pt] coordinates {(1, 0.333) (2, 0.338) (3, 0.333) (4, 0.272) (5, 0.064) (6, 0.084) (9, 0.006)}; 
                \addlegendentry{Llama 3.3}
                \addplot+[line width=1.64pt] coordinates {(1, 0.333) (2, 0.329) (3, 0.241) (4, 0.088) (5, 0.064) (6, 0.047) (9, 0.055)}; 
                \addlegendentry{DeepSeek}
                \addplot+[line width=1.64pt] coordinates {(1, 0.687) (2, 0.447) (3, 0.312) (4, 0.285) (5, 0.088) (6, 0.132) (9, 0.034)}; 
                \addlegendentry{GPT-4.1}
                \addplot+[line width=1.64pt] coordinates {(1, 0.561) (2, 0.346) (3, 0.239) (4, 0.0) (5, 0.108) (6, 0.022) (9, 0.12)}; 
                \addlegendentry{Qwen 3}
            
              \end{axis}
            \end{tikzpicture}
        }
        \caption*{(a) NL}
        \label{fig:a}
    \end{minipage}
    \begin{minipage}[h]{0.33\columnwidth}
        \centering
        \resizebox{1.0\textwidth}{!}{%
            \begin{tikzpicture}
              \begin{axis}[
                    xlabel={\# Attrs.},
                    grid=major,
                    ymin=0.0, ymax=0.75,
                    xmin=1.0, xmax=9.0,
                    legend style = {at={(0.23,1.0)}, legend columns=3, anchor=north west, fill=white,align=left},
                    width=11.1cm,
                    height=8cm,
                    cycle multiindex* list={
                      mark list*\nextlist
                      color list
                    }
                  ]

                \addplot+[line width=1.64pt] coordinates {(1, 0.069) (2, 0.105) (3, 0.191) (4, 0.0) (5, 0.005) (6, 0.0) (9, 0.0)}; 
                \addlegendentry{Mistral}
                \addplot+[line width=1.64pt] coordinates {(1, 0.165) (2, 0.18) (3, 0.303) (4, 0.029) (5, 0.002) (6, 0.0) (9, 0.0)}; 
                \addlegendentry{Qwen 2.5}
                \addplot+[line width=1.64pt] coordinates {(1, 0.317) (2, 0.339) (3, 0.117) (4, 0.117) (5, 0.022) (6, 0.002) (9, 0.021)}; 
                \addlegendentry{Llama 3.1}
                \addplot+[line width=1.64pt] coordinates {(1, 0.146) (2, 0.128) (3, 0.028) (4, 0.0) (5, 0.021) (6, 0.043) (9, 0.015)}; 
                \addlegendentry{GPT-4.1 mini}
                \addplot+[line width=1.64pt] coordinates {(1, 0.288) (2, 0.31) (3, 0.383) (4, 0.0) (5, 0.009) (6, 0.0) (9, 0.003)}; 
                \addlegendentry{Gemma 2}
                \addplot+[line width=1.64pt] coordinates {(1, 0.356) (2, 0.343) (3, 0.33) (4, 0.272) (5, 0.08) (6, 0.076) (9, 0.006)}; 
                \addlegendentry{Llama 3.3}
                \addplot+[line width=1.64pt] coordinates {(1, 0.347) (2, 0.224) (3, 0.222) (4, 0.117) (5, 0.062) (6, 0.065) (9, 0.037)}; 
                \addlegendentry{DeepSeek}
                \addplot+[line width=1.64pt] coordinates {(1, 0.261) (2, 0.269) (3, 0.07) (4, 0.0) (5, 0.022) (6, 0.087) (9, 0.012)}; 
                \addlegendentry{GPT-4.1}
                \addplot+[line width=1.64pt] coordinates {(1, 0.453) (2, 0.271) (3, 0.321) (4, 0.0) (5, 0.074) (6, 0.033) (9, 0.064)}; 
                \addlegendentry{Qwen 3}
            
              \end{axis}
            \end{tikzpicture}
        }
        \caption*{(b) SQL}
        \label{fig:b}
    \end{minipage}
    \begin{minipage}[h]{0.33\columnwidth}
        \centering
        \resizebox{1.0\textwidth}{!}{%
            \begin{tikzpicture}
              \begin{axis}[
                    xlabel={\# Attrs.},
                    grid=major,
                    ymin=0.0, ymax=0.75,
                    xmin=1.0, xmax=9.0,
                    legend style = {at={(0.23,1.0)}, legend columns=3, anchor=north west, fill=white,align=left},
                    width=11.1cm,
                    height=8cm,
                    cycle multiindex* list={
                      mark list*\nextlist
                      color list
                    }
                  ]
            
                \addplot+[line width=1.64pt] coordinates {(1, 0.278) (2, 0.191) (3, 0.35) (4, 0.088) (5, 0.002) (6, 0.0) (9, 0.0)}; 
                \addlegendentry{Mistral}
                \addplot+[line width=1.64pt] coordinates {(1, 0.322) (2, 0.112) (3, 0.321) (4, 0.029) (5, 0.001) (6, 0.0) (9, 0.004)}; 
                \addlegendentry{Qwen 2.5}
                \addplot+[line width=1.64pt] coordinates {(1, 0.355) (2, 0.442) (3, 0.378) (4, 0.088) (5, 0.061) (6, 0.0) (9, 0.01)}; 
                \addlegendentry{Llama 3.1}
                \addplot+[line width=1.64pt] coordinates {(1, 0.388) (2, 0.246) (3, 0.288) (4, 0.125) (5, 0.059) (6, 0.065) (9, 0.005)}; 
                \addlegendentry{GPT-4.1 mini}
                \addplot+[line width=1.64pt] coordinates {(1, 0.337) (2, 0.362) (3, 0.345) (4, 0.0) (5, 0.007) (6, 0.002) (9, 0.001)}; 
                \addlegendentry{Gemma 2}
                \addplot+[line width=1.64pt] coordinates {(1, 0.403) (2, 0.412) (3, 0.362) (4, 0.242) (5, 0.076) (6, 0.076) (9, 0.042)}; 
                \addlegendentry{Llama 3.3}
                \addplot+[line width=1.64pt] coordinates {(1, 0.387) (2, 0.296) (3, 0.219) (4, 0.242) (5, 0.037) (6, 0.087) (9, 0.139)}; 
                \addlegendentry{DeepSeek}
                \addplot+[line width=1.64pt] coordinates {(1, 0.541) (2, 0.29) (3, 0.399) (4, 0.25) (5, 0.132) (6, 0.13) (9, 0.003)}; 
                \addlegendentry{GPT-4.1}
                \addplot+[line width=1.64pt] coordinates {(1, 0.539) (2, 0.343) (3, 0.299) (4, 0.0) (5, 0.082) (6, 0.022) (9, 0.216)}; 
                \addlegendentry{Qwen 3}
            
              \end{axis}
            \end{tikzpicture}
        }
        \caption*{(c) CoT}
        \label{fig:c}
    \end{minipage}
\caption{Impact of the Number of attributes w.r.t. Tuple Similarity (TS) with NL, SQL, CoT strategies}
\label{fig:impactAttrsTSFull}
\end{figure}

\begin{figure}
    \begin{minipage}[h]{0.33\columnwidth}
        \centering
        \resizebox{1.0\columnwidth}{!}{%
            \begin{tikzpicture}
              \begin{axis}[
                    xlabel={\# Attrs.},
                    grid=major,
                    ymin=0.0, ymax=1.0,
                    xmin=1.0, xmax=9.0,
                    legend style = {at={(0.0,1.0)}, legend columns=4, anchor=north west, fill=white,align=left},
                    width=11.1cm,
                    height=8cm,
                    cycle multiindex* list={
                      mark list*\nextlist
                      color list
                    }
                  ]
            
                \addplot+[line width=1.64pt] coordinates {(1, 0.07) (2, 0.1) (3, 0.449) (4, 0) (5, 0.512) (6, 0.282) (9, 0.484)}; 
                \addlegendentry{Mistral}
                \addplot+[line width=1.64pt] coordinates {(1, 0.226) (2, 0.341) (3, 0.604) (4, 0.487) (5, 0.548) (6, 0.416) (9, 0.56)}; 
                \addlegendentry{Qwen 2.5}
                \addplot+[line width=1.64pt] coordinates {(1, 0.273) (2, 0.347) (3, 0.226) (4, 0.418) (5, 0.39) (6, 0.172) (9, 0.436)}; 
                \addlegendentry{Llama 3.1}
                \addplot+[line width=1.64pt] coordinates {(1, 0.2) (2, 0.187) (3, 0.264) (4, 0.319) (5, 0.365) (6, 0.264) (9, 0.641)}; 
                \addlegendentry{GPT-4o mini}
                \addplot+[line width=1.64pt] coordinates {(1, 0.33) (2, 0.49) (3, 0.655) (4, 0.443) (5, 0.571) (6, 0.463) (9, 0.746)}; 
                \addlegendentry{Gemma 2}
                \addplot+[line width=1.64pt] coordinates {(1, 0.395) (2, 0.513) (3, 0.604) (4, 0.655) (5, 0.598) (6, 0.72) (9, 0.75)}; 
                \addlegendentry{Llama 3.3}
                \addplot+[line width=1.64pt] coordinates {(1, 0.435) (2, 0.478) (3, 0.537) (4, 0.732) (5, 0.537) (6, 0.608) (9, 0.754)}; 
                \addlegendentry{Reasoning}
            
                \addplot+[line width=1.64pt]  coordinates {(1, 0.) (2,0.) (3,0.) (4,0.) (5,0.) (6,0.) (9,0.)};
                \addlegendentry{QWEN 3}
            
              \end{axis}
            \end{tikzpicture}
        }
        \caption*{(a) NL}
        \label{fig:aF1}
    \end{minipage}
    \begin{minipage}[h]{0.33\columnwidth}
        \centering
        \resizebox{1.0\textwidth}{!}{%
            \begin{tikzpicture}
              \begin{axis}[
                    xlabel={\# Attrs.},
                    grid=major,
                    ymin=0.0, ymax=1.0,
                    xmin=1.0, xmax=9.0,
                    legend style = {at={(0.0,1.0)}, legend columns=4, anchor=north west, fill=white,align=left},
                    width=11.1cm,
                    height=8cm,
                    cycle multiindex* list={
                      mark list*\nextlist
                      color list
                    }
                  ]
                \addplot+[line width=1.64pt] coordinates {(1, 0.305) (2, 0.387) (3, 0.576) (4, 0.477) (5, 0.501) (6, 0.422) (9, 0.527)}; 
                \addlegendentry{Mistral}
                \addplot+[line width=1.64pt] coordinates {(1, 0.242) (2, 0.4) (3, 0.602) (4, 0.615) (5, 0.545) (6, 0.422) (9, 0.619)}; 
                \addlegendentry{Qwen 2.5}
                \addplot+[line width=1.64pt] coordinates {(1, 0.232) (2, 0.429) (3, 0.6) (4, 0.514) (5, 0.555) (6, 0.426) (9, 0.601)}; 
                \addlegendentry{Llama 3.1}
                \addplot+[line width=1.64pt] coordinates {(1, 0.344) (2, 0.432) (3, 0.5) (4, 0.693) (5, 0.542) (6, 0.541) (9, 0.663)}; 
                \addlegendentry{GPT-4o mini}
                \addplot+[line width=1.64pt] coordinates {(1, 0.383) (2, 0.541) (3, 0.629) (4, 0.439) (5, 0.54) (6, 0.451) (9, 0.732)}; 
                \addlegendentry{Gemma 2}
                \addplot+[line width=1.64pt] coordinates {(1, 0.337) (2, 0.468) (3, 0.621) (4, 0.583) (5, 0.606) (6, 0.704) (9, 0.759)}; 
                \addlegendentry{Llama 3.3}
                \addplot+[line width=1.64pt] coordinates {(1, 0.335) (2, 0.507) (3, 0.593) (4, 0.551) (5, 0.602) (6, 0.613) (9, 0.79)}; 
                \addlegendentry{Reasoning}

                \addplot+[line width=1.64pt]  coordinates {(1, 0.) (2,0.) (3,0.) (4,0.) (5,0.) (6,0.) (9,0.)};
                \addlegendentry{QWEN 3}
            
              \end{axis}
            \end{tikzpicture}
        }
        \caption*{(b) SQL}
        \label{fig:bF1}
    \end{minipage}
    \begin{minipage}[h]{0.33\columnwidth}
        \centering
        \resizebox{1.0\textwidth}{!}{%
            \begin{tikzpicture}
              \begin{axis}[
                    xlabel={\# Attrs.},
                    grid=major,
                    ymin=0.0, ymax=1.0,
                    xmin=1.0, xmax=9.0,
                    legend style = {at={(0.0,1.0)}, legend columns=4, anchor=north west, fill=white,align=left},
                    width=11.1cm,
                    height=8cm,
                    cycle multiindex* list={
                      mark list*\nextlist
                      color list
                    }
                  ]

                \addplot+[line width=1.64pt] coordinates {(1, 0.181) (2, 0.346) (3, 0.544) (4, 0.576) (5, 0.524) (6, 0.413) (9, 0.562)}; 
                \addlegendentry{Mistral}
                \addplot+[line width=1.64pt] coordinates {(1, 0.348) (2, 0.359) (3, 0.595) (4, 0.491) (5, 0.528) (6, 0.39) (9, 0.574)}; 
                \addlegendentry{Qwen 2.5}
                \addplot+[line width=1.64pt] coordinates {(1, 0.349) (2, 0.545) (3, 0.577) (4, 0.48) (5, 0.553) (6, 0.394) (9, 0.753)}; 
                \addlegendentry{Llama 3.1}
                \addplot+[line width=1.64pt] coordinates {(1, 0.389) (2, 0.307) (3, 0.64) (4, 0.304) (5, 0.554) (6, 0.576) (9, 0.673)}; 
                \addlegendentry{GPT-4o mini}
                \addplot+[line width=1.64pt] coordinates {(1, 0.324) (2, 0.435) (3, 0.59) (4, 0.512) (5, 0.565) (6, 0.455) (9, 0.743)}; 
                \addlegendentry{Gemma 2}
                \addplot+[line width=1.64pt] coordinates {(1, 0.453) (2, 0.582) (3, 0.607) (4, 0.519) (5, 0.604) (6, 0.685) (9, 0.808)}; 
                \addlegendentry{Llama 3.3}
                \addplot+[line width=1.64pt] coordinates {(1, 0.479) (2, 0.502) (3, 0.529) (4, 0.683) (5, 0.543) (6, 0.65) (9, 0.787)}; 
                \addlegendentry{Reasoning}
            
                \addplot+[line width=1.64pt] coordinates {(1, 0.) (2,0.) (3,0.) (4,0.) (5,0.) (6,0.) (9,0.)};
                \addlegendentry{QWEN 3}
            
              \end{axis}
            \end{tikzpicture}
        }
        \caption*{(c) CoT}
        \label{fig:cF1}
    \end{minipage}
\caption{Impact of the Number of attributes w.r.t. F1 with NL, SQL, CoT strategies}
\label{fig:impactAttrsF1Full}
\end{figure}

\section{Error Analysis}

To better understand the limitations of LLMs in factual table generation, we conducted an error analysis on GPT-4.1's outputs. We randomly selected 100 examples in which all three querying strategies (NL, SQL, and CoT) produced non-perfect scores (less than 1.0) based on the average of \textsc{F1} and \textsc{TS}. Each model output was compared against the expected tupleset to identify common failure patterns. Figure~\ref{fig:error-bar-chart} reports the error distribution. Each example could belong to multiple error types.

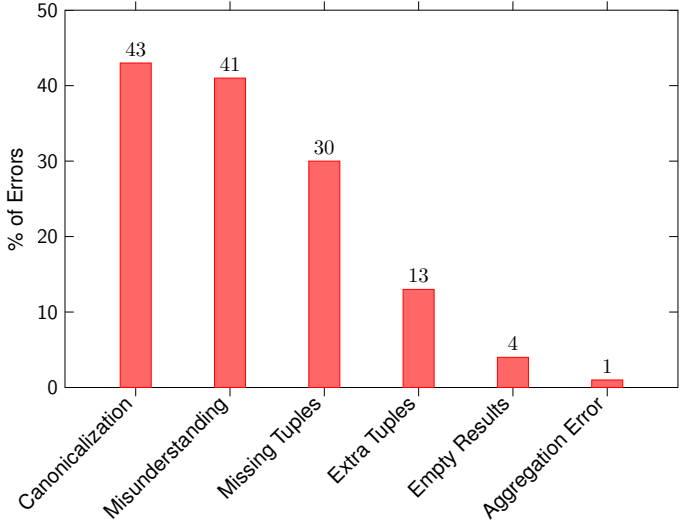
\begin{figure}[ht]
\centering
\resizebox{0.65\columnwidth}{!}{%
\begin{tikzpicture}
\begin{axis}[
    ybar,
    symbolic x coords={
        Canonicalization,
        Misunderstanding,
        Missing Tuples,
        Extra Tuples,
        Empty Results,
        Aggregation Error
    },
    xtick=data,
    x tick label style={rotate=45, anchor=east},
    ymin=0,
    ymax=50,
    ylabel={\% of Errors},
    bar width=15pt,
    width=12cm,
    height=8cm,
    enlarge x limits=0.15,
    nodes near coords,
    nodes near coords align={vertical},
    axis line style={black},
    tick style={black},
    tick label style={black},
    every node near coord/.append style={black}
]
\addplot+[draw=red, fill=red!60] coordinates {
    (Canonicalization, 43)
    (Misunderstanding, 41)
    (Missing Tuples, 30)
    (Extra Tuples, 13)
    (Empty Results, 4)
    (Aggregation Error, 1)
};
\end{axis}
\end{tikzpicture}
}
\caption{Breakdown of common error types in GPT-4.1 outputs on factual table generation. Each bar represents the percentage of 100 analyzed examples where the error type was observed.}
\label{fig:error-bar-chart}
\end{figure}

We found that \textbf{43\%} of errors come from \textbf{Canonicalization} issues, where semantically equivalent values were not matched due to limitations in our string similarity metric based on edit distance. Examples include mismatches such as ``USA'' vs. ``United States of America'' or ``s'' vs. ``s-block''. These string-based discrepancies led to false negatives but were rare in numerical values, where our proposed metric is more robust. A potential solution could involve incorporating LLMs as semantic judges to verify whether two strings refer to the same real-world entity. However, this must be done judiciously, as calling an LLM for each comparison can significantly increase evaluation time, especially for tupleset that involves a high number of cells.

Another \textbf{41\%} of errors were categorized as \textbf{Misunderstanding}, where the LLM misunderstood the intended meaning of a field. A recurring case was in the domain of chemical elements: when asked for the ``origin" of elements, models often returned the etymology of the name rather than the scientific classification. For instance, the expected answer for hydrogen's origin was ``primordial'', but the model returned ``Greek: hydro (water) and genes (forming)''. Interestingly, such errors disappeared when the attribute was used as a filter in the query's WHERE clause, for example, WHERE origin=``primordial''. This kind of error highlights the need to be clear in the values that we expect the LLM to return. For example, querying the LLM with prompts with few-shot examples could help to mitigate this kind of error. 

\textbf{30\%} of errors were due to \textbf{Missing Tuples}, where the model returned only a partial set of expected rows. These omissions were closely tied (80\% of the time) to queries with numerical conditions in the WHERE clause, particularly involving inequality operators such as $>$, $>=$, $<$, or $<=$. In contrast, equality conditions rarely led to missing results. This trend aligns with prior experimental results (Table~\ref{tab:breakdownQueryComplexity}) showing that numerical conditions are harder for LLMs to process than categorical ones.

Another \textbf{13\%} of errors were attributed to \textbf{Extra Tuples}, where the LLM returned rows not present in the expected result. These typically occurred in queries with mixed WHERE conditions (both categorical and numerical). A notable pattern emerged: in cases where the numerical condition used the = operator, the LLM often hallucinated by forcing the condition’s value into all returned rows, regardless of factual correctness. For example, when asked for tuples satisfying nationality=American AND birth\_year=1937, the model returned multiple tuples with the correct nationality but assigned 1937 as the birth year across the board, even for entities with different birth years.

Lastly, \textbf{4\%}of the cases were \textbf{Empty Results}, which can be seen as extreme cases of missing tuples, and \textbf{1\%} were due to \textbf{Aggregation Errors}, where the model failed to retrieve correct base data, resulting in incorrect aggregate computations.

Missing Tuples and Extra Tuples errors highlight a key challenge when querying an LLM: the handling of conditions in the WHERE clause. Our findings indicate that while LLMs can generally interpret categorical conditions correctly, numerical conditions often lead to hallucinations or incomplete results. 
A promising future research direction is to systematically investigate which types of conditions are reliably handled by LLMs during query execution and which are prone to errors or hallucinations. 
This analysis would help define a boundary between conditions that can be safely included in the LLM prompt and those that should instead be applied as post-processing filters, i.e., all tuples are retrieved with the LLM and the filtering is done as a separate step. Understanding this distinction could lead to more robust hybrid querying strategies that combine the generative capabilities of LLMs with traditional filtering techniques.


\end{document}